\documentclass[journal]{IEEEtran}
\usepackage{scalerel}
\usepackage{tikz}
\usetikzlibrary{svg.path}
\usepackage{makecell}
\usepackage{tabularx}
\usepackage{graphicx}
\usepackage{textcomp}
\usepackage{hyperref}
\usepackage{subcaption}
\usepackage{array}
\usepackage{framed,multirow}
\usepackage{caption}
\usepackage[T1]{fontenc}
\usepackage[latin9]{inputenc}
\usepackage{babel}
\usepackage{xcolor}
\usepackage[table]{colortbl}
\usepackage{collcell}
\usepackage{hhline}
\usepackage{pgf}
\usepackage{multirow}
\usepackage{cite}

\definecolor{orcidlogocol}{HTML}{A6CE39}
\tikzset{
  orcidlogo/.pic={
    \fill[orcidlogocol] svg{M256,128c0,70.7-57.3,128-128,128C57.3,256,0,198.7,0,128C0,57.3,57.3,0,128,0C198.7,0,256,57.3,256,128z};
    \fill[white] svg{M86.3,186.2H70.9V79.1h15.4v48.4V186.2z}
                 svg{M108.9,79.1h41.6c39.6,0,57,28.3,57,53.6c0,27.5-21.5,53.6-56.8,53.6h-41.8V79.1z M124.3,172.4h24.5c34.9,0,42.9-26.5,42.9-39.7c0-21.5-13.7-39.7-43.7-39.7h-23.7V172.4z}
                 svg{M88.7,56.8c0,5.5-4.5,10.1-10.1,10.1c-5.6,0-10.1-4.6-10.1-10.1c0-5.6,4.5-10.1,10.1-10.1C84.2,46.7,88.7,51.3,88.7,56.8z};
  }
}

\newcommand\orcid[1]{\href{https://orcid.org/#1}{\mbox{\scalerel*{
\begin{tikzpicture}[yscale=-1,transform shape]
\pic{orcidlogo};
\end{tikzpicture}
}{|}}}}

\def\colorModel{hsb}

\newcommand\ColCell[1]{
  \pgfmathparse{#1<1?1:0} 
    \ifnum\pgfmathresult=0\relax\color{white}\fi
  \pgfmathsetmacro\compA{0}      
  \pgfmathsetmacro\compB{#1/1} 
  \pgfmathsetmacro\compC{1}      
  \edef\x{\noexpand\centering\noexpand\cellcolor[\colorModel]{\compA,\compB,\compC}}\x #1
  } 
\newcolumntype{E}{>{\collectcell\ColCell}m{0.55cm}<{\endcollectcell}}  
\newcommand*\rot{\rotatebox{90}}

\def\tsc#1{\csdef{#1}{\textsc{\lowercase{#1}}\xspace}}
\tsc{WGM}
\tsc{QE}
\tsc{EP}
\tsc{PMS}
\tsc{BEC}
\tsc{DE}

\usepackage{amsmath}
\usepackage{amssymb}
\usepackage{algorithmic}
\usepackage{algorithm}
\usepackage{ragged2e}
\begin{document}

\title{Intelligent Monitoring of Stress Induced by Water Deficiency in Plants using Deep Learning}

\author{*Shiva~Azimi \orcid{0000-0001-9299-5743},
        *Rohan~Wadhawan \orcid{0000-0001-8100-668X}, ~\IEEEmembership{Member,~IEEE},
        and ~Tapan~K.~Gandhi, \orcid{0000-0002-3532-9389}, ~\IEEEmembership{Senior Member,~IEEE}
\thanks{Sh. Azimi, R. Wadhawan and T.K.Gandhi are with the Department of Electrical Engineering, Indian Institute of Technology-Delhi, New Delhi 110016, India. (* Sh. Azimi and R. Wadhawan are co-first authors, Corresponding author: T.K.Gandhi.) \\
E-mail: shiva.azimi@yahoo.com, rohanwadhawan7@gmail.com and tgandhi@ee.iitd.ac.in}
\thanks{Manuscript received July 24, 2021; revised August 21, 2021; accepted August 29, 2021.}}

\maketitle
\begin{abstract}
In the recent decade, high-throughput plant phenotyping techniques, which combine non-invasive image analysis and machine learning, have been successfully applied to identify and quantify plant health and diseases. However, these techniques usually do not consider the progressive nature of plant stress and often require images showing severe signs of stress to ensure high confidence detection, thereby reducing the feasibility for early detection and recovery of plants under stress. To overcome the problem mentioned above, we propose a deep learning pipeline for the temporal analysis of the visual changes induced in the plant due to stress and apply it to the specific water stress identification case in Chickpea plant shoot images. For this, we have considered an image dataset of two chickpea varieties JG-62 and Pusa-372, under three water stress conditions; control, young seedling, and before flowering, captured over five months. We have employed a variant of Convolutional Neural Network - Long Short Term Memory (CNN-LSTM) network to learn spatio-temporal patterns from the chickpea plant dataset and use them for water stress classification. Our model has achieved ceiling level classification performance of \textbf{98.52\%} on JG-62 and \textbf{97.78\%} on Pusa-372 chickpea plant data and has outperformed the best reported time-invariant technique by at least \textbf{14\%} for both JG-62 and Pusa-372 species, to the best of our knowledge. Furthermore, our CNN-LSTM model has demonstrated robustness to noisy input, with a less than 2.5 \% dip in average model accuracy and a small standard deviation about the mean for both species.  Lastly, we have performed an ablation study to analyze the performance of the CNN-LSTM model by decreasing the number of temporal session data used for training.

\end{abstract}



%

\section{Introduction}\label{sec:introduction}
\IEEEPARstart{I}{t} has been estimated that agricultural production should be doubled by 2050 in order to meet the demands of a growing world population. Achieving this goal poses a serious challenge to farming as the current agricultural production growth rate of $1.3 \%$ per annum is below the population growth rate. To achieve the required agricultural growth rate, we require modern agricultural practices that focus more on precision, and automated farming \cite{shadrin2019enabling}. In turn, this will employ a wide array of Internet of Things (IoT) sensors that measure soil conditions and imaging devices that keep track of specific traits such as color, size, and shape of the crops. Furthermore, we need to take a multidisciplinary approach that merges plant science, robotics, computer vision, and environmental sciences. Plant phenotyping is one such method that deals with the measurement of observable traits of a plant in reaction to genetic and environmental changes and has a large number of applications in plant science including plant breeding, quality assessments, and stress identification. Computer vision-based plant phenomics has a significant role in precision farming as it provides easy, fast, and highly automated methods for plant health and growth monitoring \cite{patricio2018computer}. Additionally, it has been used for other tasks such as determining whether a plant is a crop or a weed and the soil's chemical content using near-infrared and hyperspectral imaging.

Most manual plant phenotyping approaches are costly, time-consuming, destructive, and cumbersome, thereby necessitating the development and use of high-throughput, non-invasive, and image-based plant phenotyping techniques to identify the stress levels in plants. These methods are fast, highly automated, and more accurate. Further, image-based plant phenotyping can be conducted inside a laboratory, inside a controlled chamber room, or on the field \cite{bai2016multi}. These phenotyping techniques include two fundamental steps: the data acquisition step and the data analysis-inference step. With the recent developments in visible light, infrared and computational photography, capturing high-resolution images in both the visible and the hyperspectral has become straightforward and expedient. However, reliable and efficient data acquisition and processing methods often require expertise in biology, mathematics, and computer vision. 

Moreover, phenotyping applications usually involve the processing and analysis of a huge amount of data. Machine Learning (ML) methods have been proven to be quite efficient in the analysis of big data in research areas such as health and economics \cite{ singh2016machine}. However, the traditional ML techniques suffer from the limitation imposed by hand-crafted features. These hand-crafted features often lack generality and are unable to model complex features. This inherent limitation of the classical ML techniques has shifted the focus on Deep Learning (DL) based approaches to ML \cite{lecun2015deep}. 

One such DL architecture commonly used in various computer vision tasks is the Convolutional Neural Network (CNN) \cite{lecun1990handwritten}. CNN's possess convolutional layers for detecting visual features from images \cite{lecun1998gradient}. Further, it has been applied in several computer vision applications such as life sciences, medicine, and farming \cite{kamilaris2018deep}. It has been widely employed for classifying plants and leaves in farming \cite{lee2017deep}. It has also been used in related applications like counting the number of seeds per pod for soybeans\cite{uzal2018seed}, the number of wheat ears under field conditions \cite{madec2019ear}, plant identification \cite{sun2017deep}, identification of plant diseases \cite{barbedo2019plant}, moisture measurement of sweetcorn \cite{zhang2020development}, etc. Moreover, DL-based predictive methods have been applied in the farming domain, such as finding out future farming parameters - produce estimation \cite{kuwata2015estimating}, the soil moisture content in the field \cite{song2016genome}, prediction of the growth dynamics of plant leaves \cite{shadrin2019enabling}, and crop weather requirements \cite{sehgal2017crop}. 

This paper focuses on abiotic stresses that are caused due to external environmental factors and often adversely affect agricultural productivity. Water and nitrogen stresses are the two most crucial abiotic stresses in plants that can change plants' physiological traits. The effect of water-induced stresses, which is a consequence of excessive or inadequate watering content in the soil, inhibits photosynthesis and plants' growth. To better manage water stress and minimize crop loss resulting from it, we need to develop methods to quickly evaluate water stress without damaging the plants. 

Even though CNN has been proven very promising for image-based stress detection and classification in plants \cite{azimi2021deep}, it applies a limiting assumption of treating plant images taken at different moments in time equivalently. We know that visual changes introduced due to stress do not become discernible immediately after stress; instead, this change is progressive. However, due to CNN's time-invariant nature, it is unable to learn temporal patterns and consequently is unable to classify a stress condition with high confidence \cite{gao2020deep, azimi2021deep}. Further, the time-invariant approach also requires images showing severe signs of stress to ensure high confidence detections, thereby reducing this approach's feasibility for early detection and recovery of plants under stress. Therefore, there is a need for a technique that analyses this progressive visual change in stressed plants. This technique should classify stress with high confidence, even when available plant images do not show a sign of severe stress, as it can help us to identify stress in the plants at an early stage. 

This paper proposes a deep learning-based temporal analysis pipeline for plant water stress (water deficiency) phenotyping and demonstrates its superiority over vanilla CNN technique, which is time-invariant and only spatial. We validate the proposed approach via a detailed study that analyses changes in Chickpea plant shoot images induced due to water stress.\\
Chickpea {\it{(Cicer arietinum L.)}} is one of the crucial crops among pulses and is an excellent source of key nutrients such as proteins, iron, carbohydrates, and folic acid \cite{kumar2018combining}. Consumption of chickpea in India is the largest in the world, contributing to $75\%$ of the world's production and consumption \cite{kumar2018update}. Due to the growing concerns over food security, the demand for chickpea has been increasing in India and other developing countries. However, climate change and global warming are inducing various abiotic stresses and negatively affecting agricultural production. Among the abiotic stresses which impact chickpea production, stress-induced due to lack of water is the most significant one, causing up to $50\%$ of crop losses \cite{devasirvatham2018impact}. Water deficiency leads to specific physiological changes in chickpea plants such as dryness, yellow leaves, early flowering, and the reduction of leaf size and biomass \cite{gupta2018effect}. Owing to chickpea's potential towards ensuring food security in developing countries like India, it is imperative to develop image-based analysis methods for easy and early detection of water-related stress.

With this objective in mind, we make the following contributions in this paper: 
\begin{itemize}
  
\item As there are no publicly available plant shoot image datasets of pulses that can be used to detect and classify moisture-related stress conditions, we have created a dataset of Chickpea plant shoot images for the experiments proposed in this article. The dataset comprises two varieties of chickpea plant species - JG-62 and Pusa-372. 

\item We have proposed an end-to-end deep learning pipeline for identifying water stress in Chickpea plants. This pipeline employs a variant of Convolutional Neural Network - Long Short Term Memory (CNN-LSTM) to learn spatio-temporal patterns from the chickpea plant dataset and use them for water stress classification. The CNN-LSTM has achieved ceiling level classification performance of \textbf{98.52\%} on JG-62 and \textbf{97.78\%} on Pusa-372 and the chickpea plant data.

 \item  We have conducted a comparative analysis of the proposed temporal technique with CNN techniques to classify water stress in Chickpea plants. Our proposed technique outperforms the best reported CNN technique by at least 14\% for both JG-62 and Pusa-372 species.

\item We tested the robustness of our CNN-LSTM model to noisy input. Across both species, the average model accuracy dipped by less than 2.5 \%, with a small standard deviation. This ensures high and consistent classification capabilities even in noisy conditions.

\item We have performed an ablation study on the CNN-LSTM model by decreasing the number of temporal session data used for training. 
   
\end{itemize}

The rest of this article is structured as follows. The dataset and DL techniques are presented in section 2. The results are presented in section 3. Discussion on the results and application scope is provided in section 4. Finally, conclusions are provided in section 5.

\begin{figure}[b]
    \centering
    \begin{subfigure}[b]{0.21\textwidth}
        \includegraphics[height=3cm, width=4cm]{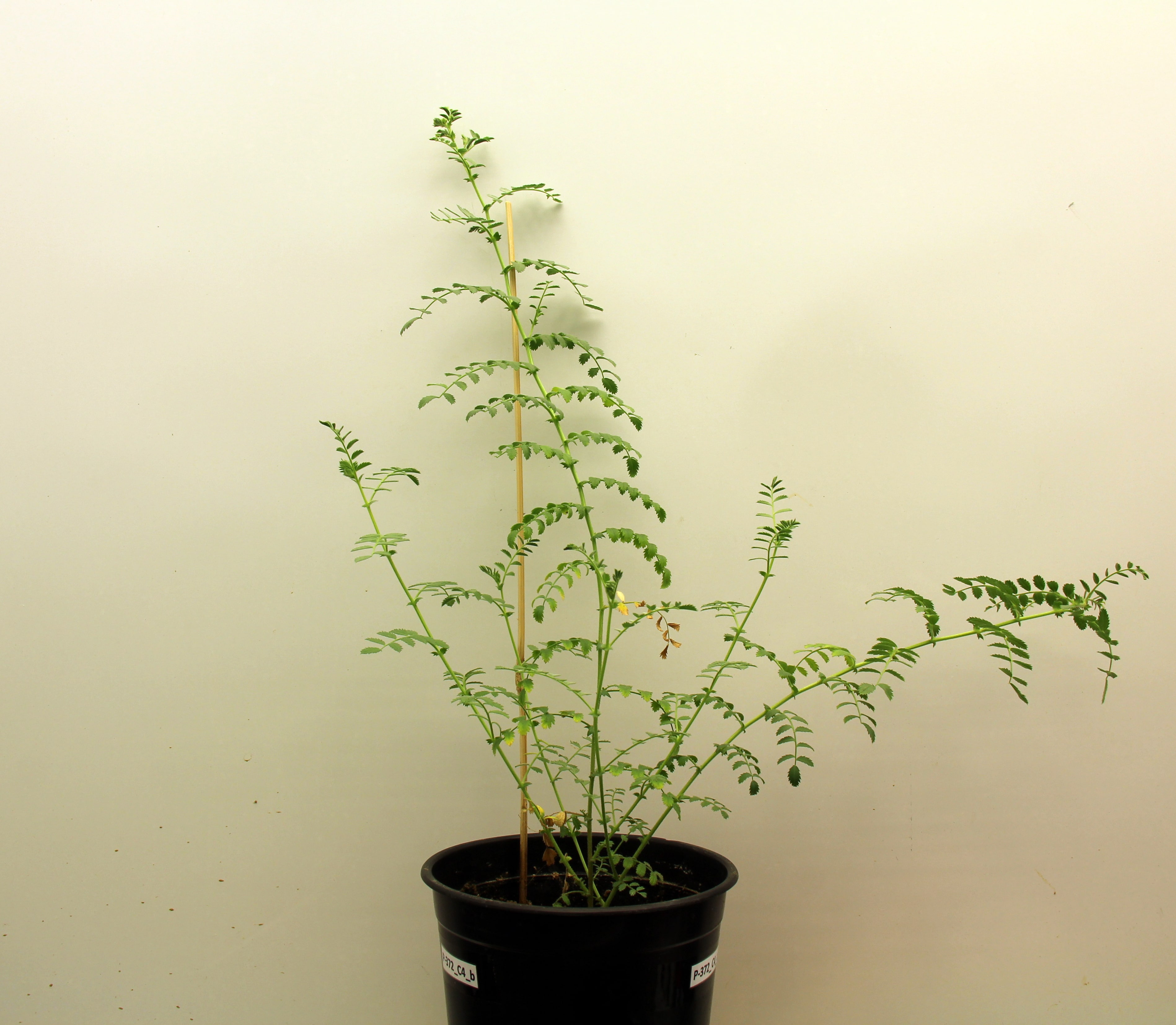}
             \label{fig1a}
    \end{subfigure}
    \quad
    \begin{subfigure}[b]{0.21\textwidth}
        \includegraphics[height=3cm, width=4cm]{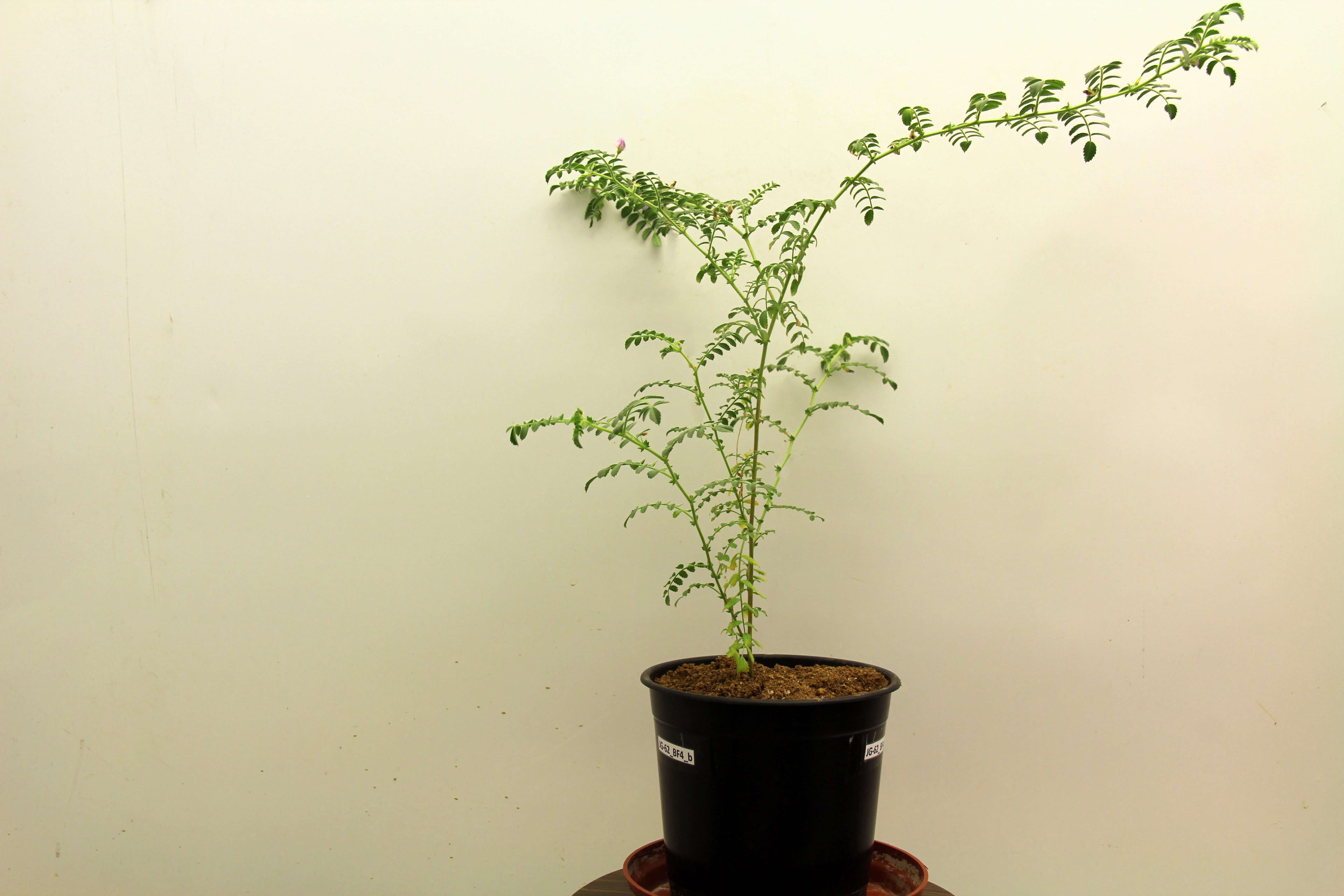}
        \label{fig1b}
    \end{subfigure}
    \caption{Visualization of replicates in stress-tolerant Pusa-372 (left) and stress-sensitive JG-62 (right) varieties from our dataset.}
\end{figure}
\section{Materials and Methods}
\begin{figure*}[h]
\centering
\includegraphics[width=14.5cm,height=8.3cm]{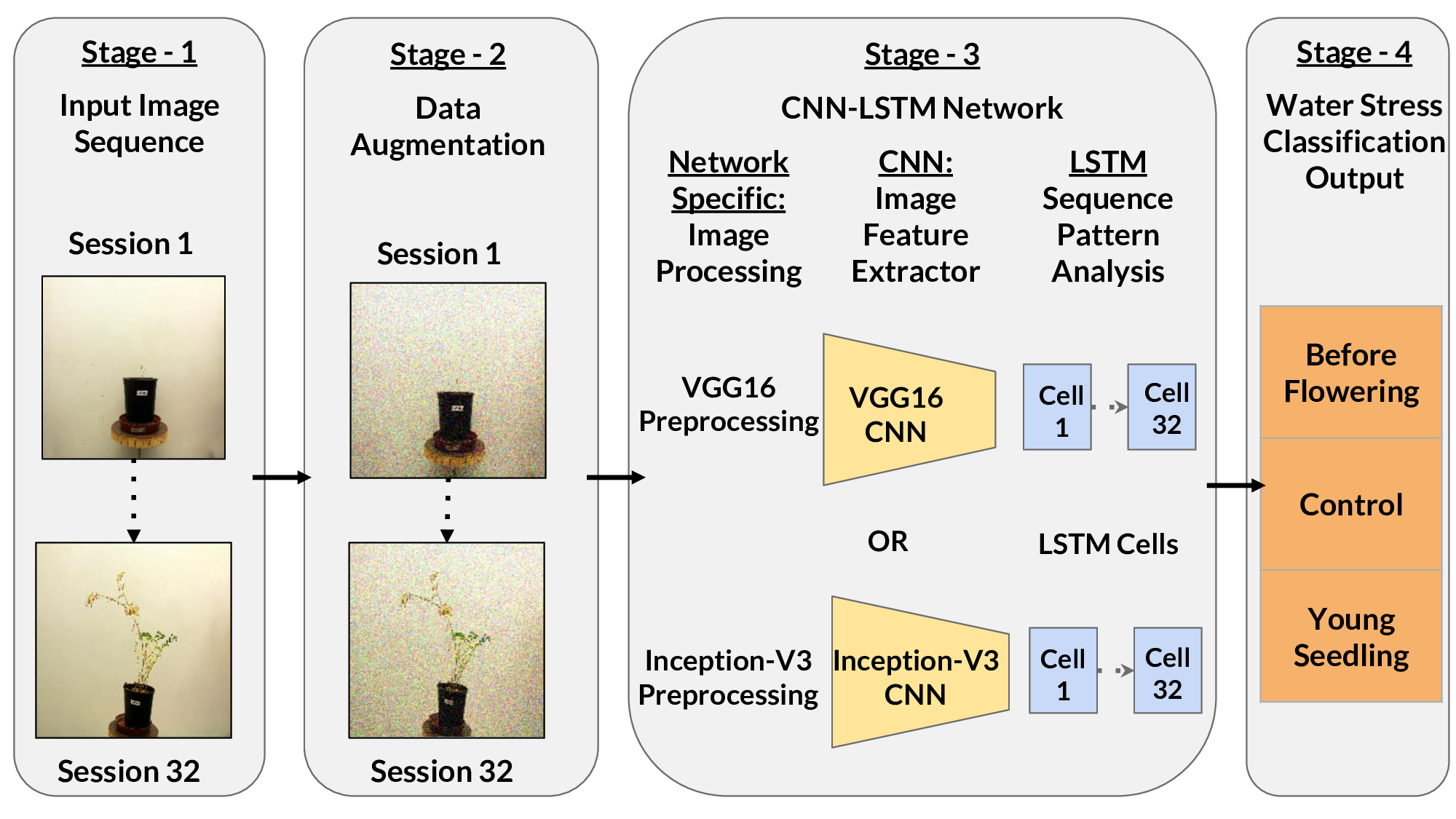}
\caption{The Four Stages of our Deep Learning Water Stress Classification Pipeline for Chickpea plant shoot images. Here, JG-62 plant images have been used to demonstrate the pipeline; noise augmentation has been shown as an example of data augmentation used.}
\label{fig:pipeline_diagram}
\end{figure*}

In this section, we describe the dataset and methodology that we use for water stress identification in chickpea plants. First, we explain our chickpea plant shoot dataset, and then we discuss the DL techniques used in this paper. Our deep learning water stress classification pipeline consists of four main stages: input, data augmentation, CNN-LSTM network, and classification output, as shown in Fig.~\ref{fig:pipeline_diagram}. These four stages are described in detail in the following subsections.

\begin{table*}[ht!]
\caption{Parameters of Chickpea plant shoot images dataset.}
\begin{center}
\scalebox{0.85}{

\begin{tabular}{|c | c | c | c | c | c |c | c | c | c | }
    \hline
    \hline
    Chickpea Variety & Light used & {\makecell{Distance of \\ camera}} & Camera & Image Type & {\makecell{Image size \\ in pixels }}& Total Images& Condition & Image labelling & No of images\\

    \hline
   \multirow{3}{*}{ $Pusa-372$}& \multirow{3}{*}{\makecell{Fluorescent \\ Tubes}}&  \multirow{3}{*}{1.5 meter} & \multirow{3}{*}{Canon EOS 60D} & \multirow{3}{*}{RGB (JPEG)} & \multirow{3}{*}{5184*3456} & \multirow{3}{*}{3840} & {\makecell{Before Flowering}} & BF & 1280\\
     \cline{8-10}
      &&&&&&& Young Seedling & YS & 1280 \\
     \cline{8-10}
       &&&&&&& Control & C & 1280 \\
   
\hline
\hline
   \multirow{3}{*}{ $JG-62$}& \multirow{3}{*}{\makecell{Fluorescent \\ Tubes}} &  \multirow{3}{*}{1.5 meter} & \multirow{3}{*}{Canon EOS 60D} & \multirow{3}{*}{RGB (JPEG)} & \multirow{3}{*}{5184*3456}  &\multirow{3}{*}{3840} & {\makecell{Before Flowering}} & BF & 1280\\
     \cline{8-10}
&&&&&&& Young Seedling & YS& 1280 \\
     \cline{8-10}
       &&&&&&& Control & C & 1280 \\
\hline
\hline
\end{tabular}}
\end{center}
\label{tab:data}
\end{table*}

\subsection{Dataset} 
Most publicly available datasets for plant health analysis only contain images on plant leaves, which is significantly less informative than the entire plant shoot image. These datasets usually show plants under biotic stress, with very few covering plants under abiotic stress. Phenotyping using complete plant shoot images offers certain advantages. Firstly, plant shoot contains more information than individual plant organs, like leaves, branches, flowers, and provide a holistic view of the plant. Secondly, capturing shoot images of a plant is faster, more robust, and provides equal or more visual features than capturing images of individual plant organs of the same plant. Thirdly, temporal analysis of shoot images over time will require low complexity models compared to the integrated temporal analysis of various plant organs, making the former more viable for real-time use. Lastly, this technique is non-destructive and non-evasive, enabling us to make observations while the plant is growing. Thus, using complete shoot images for phenotyping applications is desirable. Furthermore, to the best of our knowledge, there are no publicly available plant shoot image datasets of pulses, especially chickpea, to detect and classify moisture-related abiotic stress conditions. To this end, we created a new dataset of chickpea plant shoot images in the visible spectrum of light.

Two varieties of chickpea strains, namely - stress-tolerant Pusa-372 and stress-sensitive JG-62 - were grown in individual plant pots in the control chamber room and observed over a period of five months for this experiment. From now on, we will refer to JG-62 as JG and Pusa-372 as Pusa. The experiment was conducted in collaboration with plant scientists at the National Institute of Plant Genome Research (NIPGR). For both the varieties, plants were subjected to three different watering conditions based on the water stress applied to them. The three watering conditions are Young Seedling (YS), in which a plant was not watered for 1 week after it was 2 weeks old; Before Flowering (BF), in which a plant was not watered for 1 week after it was 5 weeks old; Control (C), in which a plant was watered throughout. Water stress changes the physical structure of plants, such as shape and color. It also reduces plant height, plant biomass, and the number of branches, leaves, and fruits in chickpea plants. We had 15 pots per species and 5 per water stress category for our experiment. The plant shoot images were captured in regular sessions at a particular time once every three days. For each pot, we have 32 sessions of data. During each image capturing session, images were taken from eight different angles, at every $45^o$. Further, the lighting condition, the camera distance, and other dataset parameters shown in the Table \ref{tab:data} below were kept the same for all plant pots (with acceptable and negligible margin of human error). Thus, in every session, we have captured 240 images across all the pots of both varieties. Overall, this dataset has a total of $7680$ images. The black pot and the white background were seen in the image. Segmentation can be applied to extract the plant shoot portion from the image but at an additional computational cost in terms of time and resources. As DL techniques are able to avoid such invariant features existing in the context, we do not apply plant shoot segmentation in our paper favoring real-time deployment over high classification accuracies. Fig. 1 shows two sample images from our dataset.

\subsection{Deep Learning Approach}
Over the years, DL techniques like CNNs have become state of the art for image classification. In our paper \cite{azimi2020water}, we employed a 23-layered custom CNN, and ResNet-18 \cite{he2016deep} model for water stress classification from chickpea plant shoot images. The ResNet-18 classifier was able to achieve 84\% and 86\% accuracy on Pusa and JG, respectively. However, this approach enforced a simplifying assumption on the dataset by treating all images belonging to one class equivalent even if they were taken at different times. Furthermore, water stress is introduced after 2 weeks, due to which the images up to that point across all the three conditions are similar to one another, thereby adding noise to the dataset. Despite this noise, the CNN classifier is robust enough to analyze water-stressed plants' patterns accurately. However, we hypothesize that time-series analysis of the plant shoots' visual features will remove this noise and produce better results.

Recurrent Neural Network (RNN) has been employed for sequential learning tasks. Long Short Term Memory (LSTM) network is an improvement over the RNN architecture \cite{hochreiter1997long}. Unlike RNN, LSTM can learn long-term dependencies and preserve useful temporal information for an extended period. They have become a state-of-the-art technique for sequence learning problems like time series analysis \cite{ma2020unauthorized}. Moreover, LSTM and CNN combined have also been successfully used in tasks requiring sequence learning of visual features \cite{bao2020cnn}, like video classification and activity recognition in videos\cite{lstm-vid1, lstm-vid2}. Our task shares similarities with activity classification in videos that predicts which activity is being performed by analyzing visual changes over time. Similarly, we need to ascertain temporal patterns resulting from visual changes induced in the chickpea plant's shoot due to water stress. CNN-LSTM architecture combines LSTM and CNN for spatio-temporal learning. Thus, we introduce a variant of the CNN-LSTM to predict water stress in chickpea plants. In this architecture, CNN pre-trained on ImageNet data extracts visual features from the chickpea plant shoot images. Then, the LSTM analyses these features over time to predict the plant's water stress condition. We also compare our previous time-invariant approach for water stress classification in chickpea plants \cite{azimi2020water} with our proposed temporal approach. Several CNN architectures have been developed over time. In this paper, we have used VGG16 \cite{vgg16} and Inception-V3 \cite{inceptionv3} architectures. Firstly, we fine-tune models of these architectures pre-trained on the ImageNet dataset \cite{deng2009imagenet} and use them for time-invariant classification of chickpea plants under water stress conditions. Secondly, we use these models as feature extractors for the CNN-LSTM models.

\textbf{VGG16:} VGG16 architecture has achieved state-of-art accuracy for image classification on the ImageNet dataset in the past. This architecture introduces the concept of stacking smaller convolutional kernels to produce an effective receptive field. This technique also decreased the total number of parameters to achieve the same receptive field and increased non-linearity due to activation across multiple stacked layers. This model was deeper and less wide than GoogLeNet (Inception-V1) \cite{inceptionv1} proposed around the same time. Although VGG16 performs better than GoogLeNet on the ImageNet dataset, it is more computationally complex and has a more significant computation, memory, and storage requirement.

\textbf{Inception-V3:} Inception-V3 architecture proposed as an improvement over its predecessors (Inception-V1 and Inception-V2) has achieved state-of-the-art accuracy for image classification on ImageNet dataset in the past. Some of the essential features of this model are: it is deeper, avoids representational bottlenecks, especially early in the network, maintains higher dimensional representation, spatial aggregation on the lower dimension, balance width, and depth of the network. In addition, it further reduces the computational complexity, both in terms of the number of parameters and cost of resources (memory and storage) compared to Inception-V1 and Inception-V2 architectures, and increases classification accuracy. As a result, Inception-V3 performs better than VGG16 on the ImageNet dataset. 

We describe the architectures, input processing, and neural network training in the subsequent sections.

\subsubsection{CNN Architecture}
\begin{figure}[t]
\includegraphics[width=\columnwidth, height = 5cm]{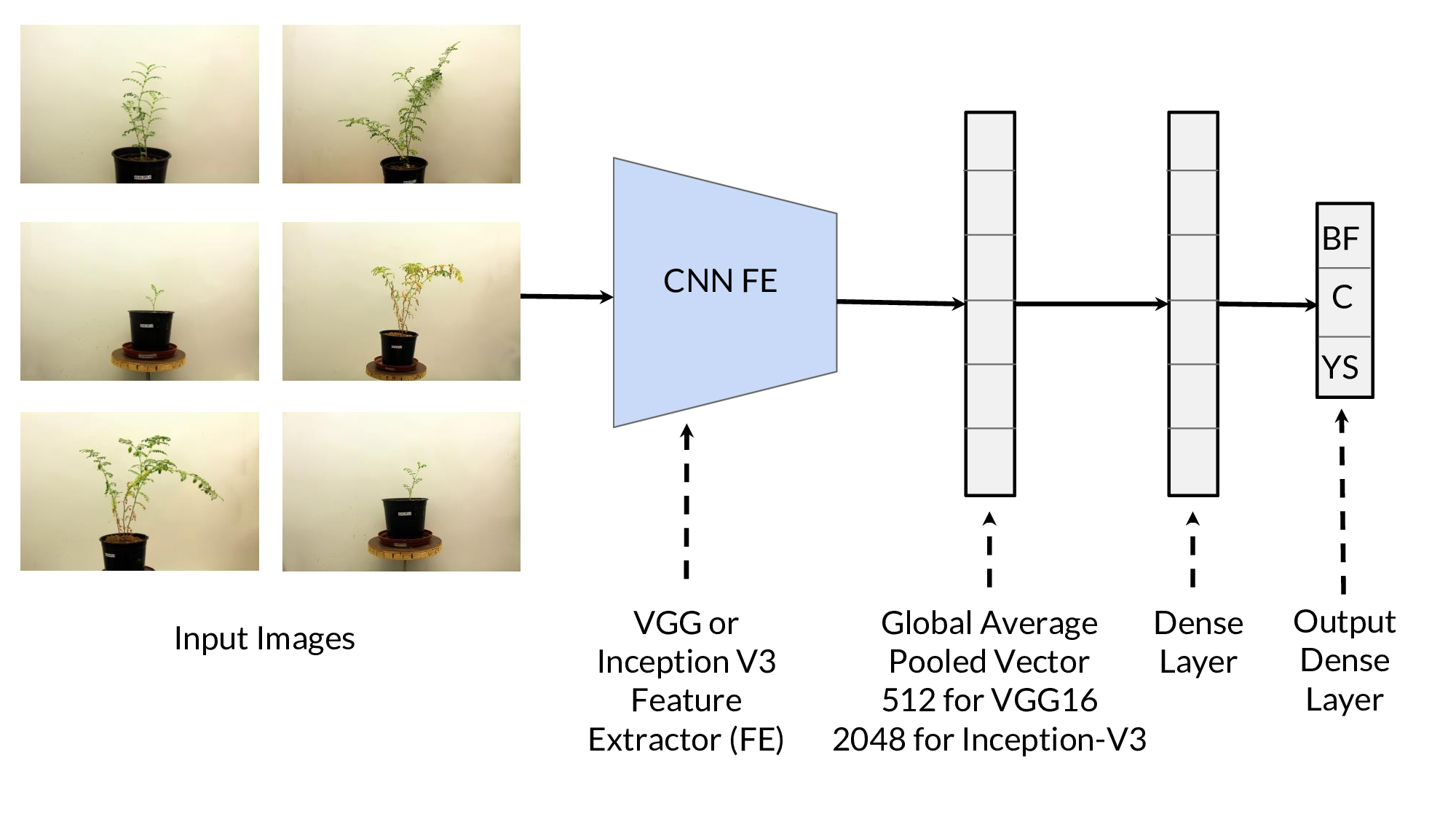}
\caption{CNN architecture used for water stress classification (BF, YS, C) in chickpea plants shoot images. In this diagram, we have used images of JG-62 chickpea species.}
\label{fig:tin_cnn_arch}
\end{figure}

This network performs a time-invariant classification analysis to identify water stress in chickpea plant shoot images. For this purpose, we use the convolutional base of the VGG16 and Inception-V3 network and remove the corresponding dense layers. Then, we perform Global Average Pooling \cite{globalavgpool} after the last Max Pooling layer. Global Average Pooling is preferred over fully connected layers for flattening the feature maps to a linear vector because it is more native to the convolution structure and enforces correspondences between feature maps and categories. Further, this layer has no parameters to optimize, which reduces the chances of over-fitting and is also more robust to the input's spatial translation. Finally, we add two dense layers after global average pooling, the first one has 512 dimensions, and the following is the output layer with three dimensions, equal to the number of classes, as shown in Fig. \ref{fig:tin_cnn_arch}. We initialize each dense layer using the Glorot uniform initializer \cite{tr-glorot} and use Softmax activation in the final output Dense layer (Equation \ref{eq:softmax}).
\begin{figure*}[t]
\centering
\includegraphics[width=\linewidth]{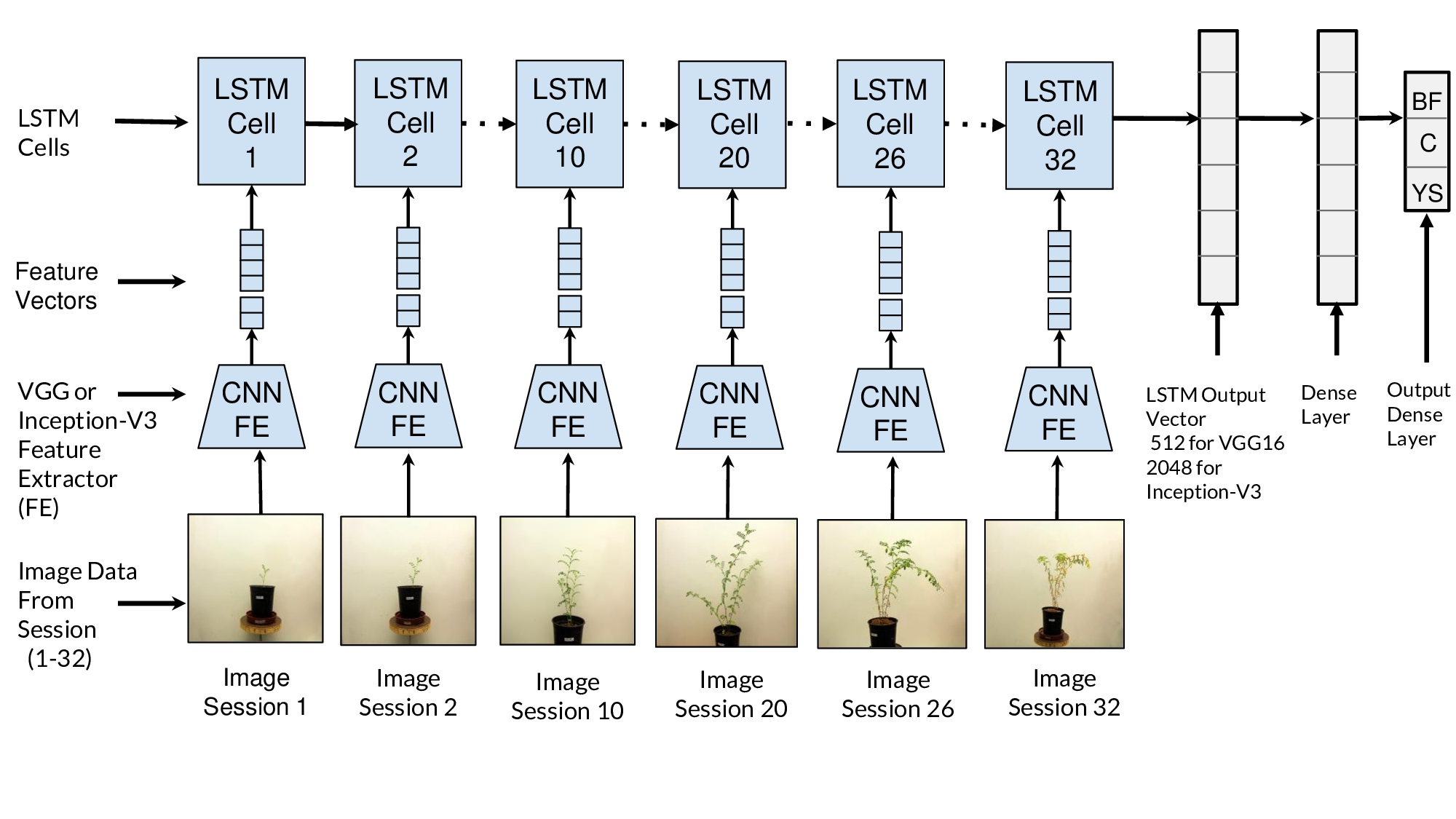}
\caption{CNN-LSTM architecture used for predicting water stress classification (BF,YS,C) in Chickpea plants. Number of LSTM cells equals the number of session data used. In this diagram, the images of a JG-62 plant sample over 32 sessions are used.}
\label{fig:lstm_cnn_arch}
\end{figure*}

\subsubsection{CNN-LSTM Architecture}
Our CNN-LSTM architecture consists of two main parts: CNN image feature extractor and LSTM to predict water stress category from the extracted features. The architecture is shown in Fig. \ref{fig:lstm_cnn_arch}.

\textbf{CNN feature extractor:} We use VGG16 and Inception-V3 models pre-trained on the ImageNet dataset to extract visual features from chickpea plant images. We employ two different feature pre-trained extractors to determine if our approach is CNN architecture-dependent. In both the models, we remove the dense layers and apply Global Average Pooling after the final Max-Pooling layer to obtain 1D vectors of size 512 and 2048 for VGG16 and Inception-V3, respectively. We use the CNN network in time distributed form, that is, the same network is shared across all time steps of subsequent LSTM network. The unrolled version is shown in Fig. \ref{fig:lstm_cnn_arch}. 

\textbf{LSTM predictor:} In our LSTM network, the number of sequentially connected cells is equal to the number of session data used for prediction, as shown in Fig. \ref{fig:lstm_cnn_arch}. This variable length of the LSTM network helps us analyze the effect of the number of data sessions on the prediction performance metrics - Accuracy, Macro Sensitivity, Macro Specificity, and Macro Precision. An ablation study on the same is reported in section \ref{sec:abstudy}. The LSTM network output is fed into a Dense Layer of size 512 dimension, which is connected to the Dense output layer of size 3, equal to the number of water stress categories. We initialize each dense layer using the Glorot uniform initializer and use Softmax activation in the final output Dense layer (Equation \ref{eq:softmax}). 

Let us mathematically trace how our proposed network processes an input image sequence of plant shoot images.
An image of an input sequence can be written as
$i_t$ defined as an image at timestep t, $i_t \in R^{m \times m}$, where the image is of dimension $m \times m,\ m = 224 $.
We define one entire image input sequence as
\[
I = \{i_t |\; Image\; at\; timestep\; t,\; t \in \mathbb{N},\: 1\leq t \leq 32,\: i_t \in R^{m \times m} \} 
\]
where $I \in R^{T  \times m  \times m}$, T is the number of time steps. Then, we chose either VGG16 or Inception-V3 CNN to extract features from images. We apply the chosen CNN feature extractor to each image of a sequence in a time-distributed manner, such that its weights  $W_c$ remain the same for all LSTM timesteps and obtain corresponding features. One feature of the output feature sequence can be written as
$x_t$ feature at timestep t, $x_t \in R^d$, where d is the size of the feature vector.
We define one entire feature output sequence as
\[
X = \{x_t |\; Feature\; at\; timestep\;  t,\; t \in \mathbb{N },\; 1\leq t \leq 32,\; x_t  \in R^d \}
\]
, where $F \in R^{T \times d}$, T is the number of timesteps. Therefore, the convolutional feature extractor simulates a function
\[ 
g  \colon  I \to X 
\]
\begin{IEEEeqnarray}{C}
\left(x_{t_{1}}, \ldots, x_{t_{T}}\right)=\left(g\left(i_{t_{1}}\right), \ldots, g\left(i_{t_{T}}\right)\right)
\label{eq1}
\end{IEEEeqnarray}
\begin{figure}[!h]
    \centering
        \includegraphics[width=\columnwidth, height = 4.2cm]{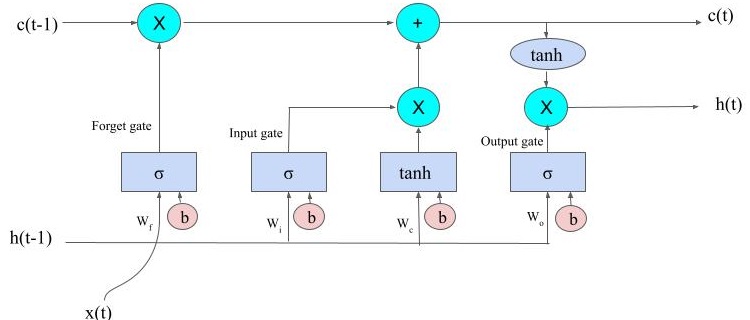}
        \label{fig:lstm}
    \caption{A Long Short-Term Memory (LSTM) cell.} 
\end{figure}

Then, we feed the feature sequence to a sequence of LSTM units. An LSTM unit comprises a cell, an input gate, an output gate, and a forget gate, as shown in Fig. 5. The cell remembers values over arbitrary time intervals, and the three gates regulate the flow of information into and out of the cell. The equations for an LSTM are defined as:

\begin{IEEEeqnarray}{C}
f_t = {\sigma_g\left(W_f x_t+U_f h_{t-1}+b_f\right)}
\label{eq2}
\end{IEEEeqnarray}

\begin{IEEEeqnarray}{C}
i_t = {\sigma_g\left(W_i x_t+U_i h_{t-1}+b_i\right)}
\label{eq3}
\end{IEEEeqnarray}

\begin{IEEEeqnarray}{C}
o_t = {\sigma_g\left(W_o x_t+U_o h_{t-1}+b_o\right)}
\label{eq4}
\end{IEEEeqnarray}

\begin{IEEEeqnarray}{C}
\Tilde{c}_t = {\sigma_c\left(W_c x_t+U_ch_{t-1}+b_c\right)}
\label{eq5}
\end{IEEEeqnarray}

\begin{IEEEeqnarray}{C}
c_t = {f_t \odot c_{t-1}+i_t \odot \Tilde{c}_t}
\label{eq6}
\end{IEEEeqnarray}

\begin{IEEEeqnarray}{C}
h_t = {o_t \odot \sigma_h(c_t)}
\label{eq7}
\end{IEEEeqnarray}

\begin{math} 
 \! x_t \in \mathbb{R}^d: Input\ feature\ vector\ to\ the\ LSTM\ unit \\
 f_t \in \mathbb{R}^h: Forget\ gate's\ activation\ vector, \\ 
 i_t \in \mathbb{R}^h: Input\ gate's\ activation\ vector, \\
 o_t \in \mathbb{R}^h: Output\ gate's\ activation\ vector, \\
 h_t \in \mathbb{R}^h: Hidden\ state\ output\ vector\ of\ the\ LSTM\ unit, \\
\tilde{c}_t \in \mathbb{R}^h: Cell\ input \ activation\ vector, \\
c_t \in \mathbb{R}^h: Cell \ state\ vector, \\
W \in \mathbb{R}^{h  \times d}: Weight\ Matrix\\ 
U \in \mathbb{R}^{h \times h} Weight\ Matrix\\ 
\ b \in \mathbb{R}^h: Bias\ Matrix \\
\sigma_g: Sigmoid\ function\\
\sigma_h: Hyperbolic\ tangent\ function \\
\end{math}

Here, d and h denote input feature and hidden state dimensions, respectively. In addition, $\odot$ denotes element-wise multiplication. Weight and bias matrices are learned during training. 

Then, we take the hidden vector, also known as output vector $h_t$ of the final LSTM unit $h_{t=T}$ and feed it as input to the classification block consisting of two dense layers. 

For the classification block, the Input is $H = h_{t=T}$, and output is $P \in  R^C$, where C is the number of classes, here $C = 3 $, $Classes = \{Before \ Flowering, Control, Young\ Seedling\}$. Then, the dense layer simulates a function \[
j  \colon  H \to P 
\] \begin{IEEEeqnarray}{C}
\left(p_{1}, p_{2}, p_{3}\right)=j\left(h_{t_{T}}\right)
\label{eq8}
\end{IEEEeqnarray}
The proposed network's output is equal to the classification block's output, that is, P. As it is a case of multi-class classification, softmax activation is applied to the output of the final fully connected layer, also called the classification layer. It helps convert the output score corresponding to each class into a probability value between 0 and 1. 

\begin{IEEEeqnarray}{C}
\label{eq:softmax}
\textnormal{Softmax}\left(p_i\right)=\frac{\exp^{p_i}}{\sum \limits_{j=1 }^{C}\exp^{p_j}}.
\end{IEEEeqnarray}
where $p_i$ is the predicted probability of a class represented as an element of the 3 dimensional output vector P.

\begin{figure*}[!h]
	\begin{minipage}{.5\textwidth}
	    \centering  
        \begin{subfigure}[t]{.2\textwidth}
            \centering
    		\includegraphics[width=2cm,height=2.25cm]{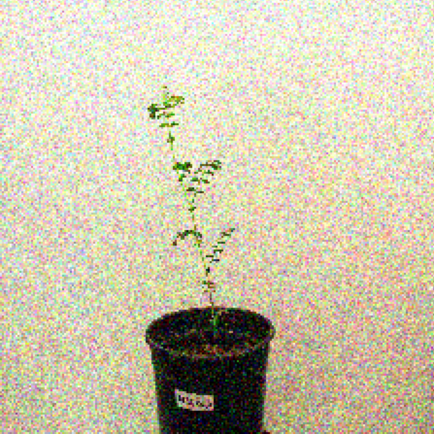}
	    \end{subfigure}
	    \hspace{0.15cm}
	    \begin{subfigure}[t]{.2\textwidth}
            \centering
            \includegraphics[width=2cm,height=2.25cm]{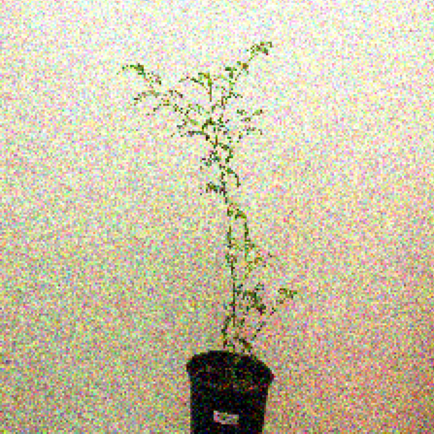}
	    \end{subfigure}
	    \hspace{0.15cm}
	    \begin{subfigure}[t]{.2\textwidth}
        \centering
            \includegraphics[width=2cm,height=2.25cm]{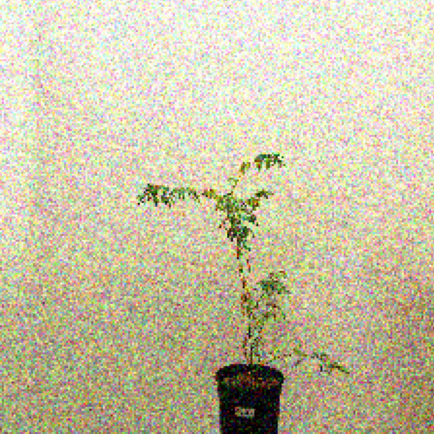}
	    \end{subfigure}%
	    \hspace{0.25cm}
	    \begin{subfigure}[t]{.2\textwidth}
            \centering
            \includegraphics[width=2cm,height=2.25cm]{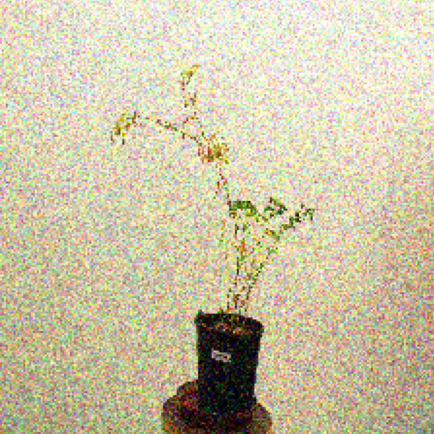}
    	\end{subfigure}%
    	\caption{Gaussian noise added to images of a given JG-62 image sequence.}
    	\label{fig:jg_noise}
	\end{minipage}
	\hspace{0.5cm}
	\begin{minipage}{.5\textwidth}
		\centering  
		\begin{subfigure}[t]{.2\textwidth}
            \centering
    		\includegraphics[width=2cm,height=2.25cm]{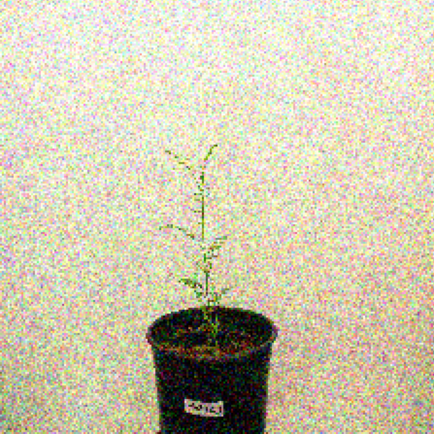}
	    \end{subfigure}
	    \hspace{0.15cm}
	    \begin{subfigure}[t]{.2\textwidth}
            \centering
            \includegraphics[width=2cm,height=2.25cm]{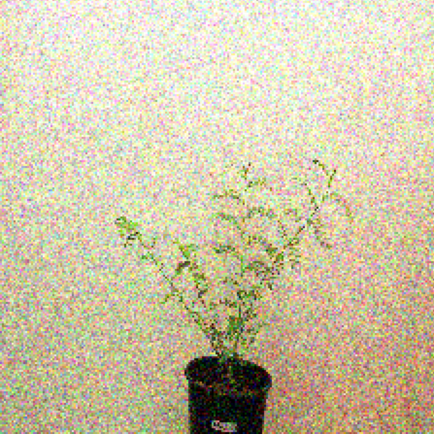}
	    \end{subfigure}
	    \hspace{0.15cm}
	    \begin{subfigure}[t]{.2\textwidth}
        \centering
            \includegraphics[width=2cm,height=2.25cm]{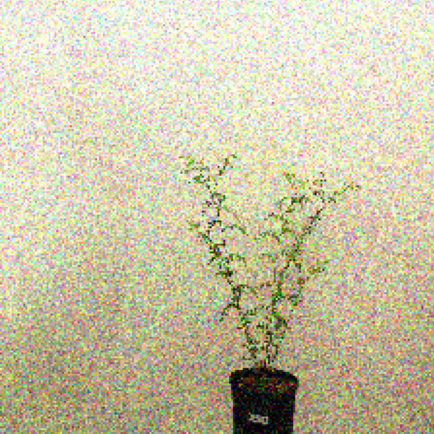}
	    \end{subfigure}%
	    \hspace{0.25cm}
	    \begin{subfigure}[t]{.2\textwidth}
            \centering
            \includegraphics[width=2cm,height=2.25cm]{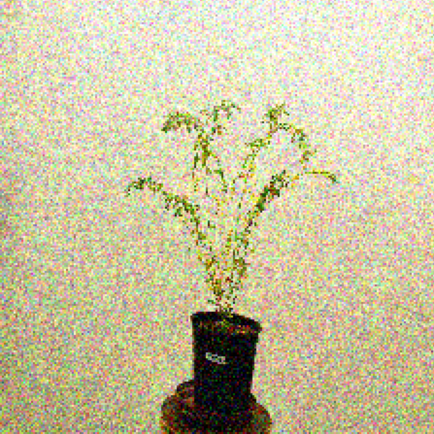}
    	\end{subfigure}%
    	\caption{Gaussian noise added to images of a given Pusa-372 image sequence.}
    	\label{fig:pusa_noise}
	\end{minipage}
\end{figure*}

\subsubsection{Input Processing}
\textbf{CNN-LSTM Network:} This section describes input dataset preparation for the CNN-LSTM Network. Each input data sequence consists of 32 images of a plant pot, one from every photo session. To ensure the robustness of classification, we consider photographs at all angles, such that all images of one data sequence have been taken from the same angle. Thus, for both JG and Pusa, we have 120 data samples each, in which there are 40 samples for each water stress category. We use RGB input images of size (224,224,3) and perform CNN network-specific image preprocessing on them before feeding them to our CNN-LSTM. While training the models, we perform data augmentation like horizontal flipping, rotation, shear, and translation to increase the training data's size on the fly. We ensure that a linear transformation is performed equivalently for each image in the image sequence. Besides linear transformations, we randomly introduce Gaussian noise perturbations to a few training samples. A typical image noise model is Gaussian, additive, independent at each pixel, and independent of the signal intensity. Further, Gaussian noise in digital images, usually a consequence of sensor noise, arises during acquisition. As data acquisition in real-world settings will often be accompanied by noise, we perform noise data augmentation to train a robust model. We perform this augmentation by sampling noise intensities from a Gaussian noise $N$ distribution, which has a mean $\mu = 0$ and a standard deviation $\sigma =15\% $ of maximum pixel intensity of any image in our dataset. \[
N \sim \mathcal{N}(0,\,\sigma^{2})
\label{eq:gauss}
\] The value of $ \sigma $ has been empirically chosen to provide the best robustness capability for noisy shoot images without making noise, a relevant feature for the model to learn. JG and Pusa noisy input samples are shown in Fig. \ref{fig:jg_noise} and \ref{fig:pusa_noise}.

\textbf{CNN Network:} This section describes input data preparation for fine-tuning pre-trained VGG16 and Inception-V3 CNNs. In this case, we equivalently treat all images of a given plant taken at different points in time. Thus, we have 1280 images per category and 3840 in total for each species. We use RGB input images of size (224,224,3) and perform network-specific image preprocessing before feeding them to the corresponding CNN for fine-tuning. Similar to the CNN-LSTM, we perform data augmentation like horizontal flipping, rotation, shear, and translation to increase the training data's size.

\subsubsection{Training}
\label{sec:training}

We optimize the CNN-LSTM network (temporal analysis) and CNN networks (time-invariant analysis) by minimizing the categorical cross-entropy loss for water stress classification. 
\begin{IEEEeqnarray}{C}\label{eq:ce}
Categorical \ Cross \ Entropy = -\sum_{i=1}^{C}y_ilog(\hat{y_i})
\end{IEEEeqnarray}
Where C is the number of classes, \(Classes = \{Before \ Flowering, Control, Young\ Seedling\}\), $y_i$ is the true class,  and $\hat{y_i}$ is predicted class, which is obtained after softmax activation, refer to Equation \eqref{eq:softmax}.

For our CNN-LSTM, we freeze the weights of the CNN and train the LSTM and the dense layer. To train this network, we backpropagate the loss and update the weights of the network using the Backpropagation Through Time \cite{bptt} algorithm. The LSTM is trained on 32 sessions data and has about 2.4M and 35M trainable parameters with VGG16 and Inception-V3 feature extractors, respectively. On the other hand, to simulate time-invariant classification, we fine-tune the pre-trained VGG16 and Inception-V3 networks on our dataset using the Backpropagation technique \cite{tr-backprop}. 

The training is performed using a mini-batch size of 32 images and neural network weights are optimized using Adam optimizer \cite{tr-adam} with learning rate \(\alpha = 0.0001\) and the other optimizer parameters being \( \{\beta_{1} = 0.9, \beta_{2} = 0.999, \epsilon = 10^{-7}\}\). Further, we train each model for 200 epochs and use them for metric evaluation.

\textbf{Training Environment}
We use Tensorflow and Keras DL framework to train our models and train them on a single Nvidia Tesla K80 GPU.

\subsubsection{Evaluation Protocol}
In this paper, we perform 5-fold stratified cross-validation for each model type (plant-variety and CNN pair). In other words, we divide the entire dataset into 5 equivalent subsets and train a model on 4 out of 5 of them. Then, we test on the remaining subset such that each subset acts as a test set once. Finally, we report the average scores across all 5 models for each performance metric - Accuracy, Macro-Sensitivity, Macro-Specificity, and Macro-Precision. We also repeat the cross-validation process 10 times to ensure robustness of the reported scores.

\subsection{Performance Evaluation Metrics}
The performance of the proposed model is evaluated using the performance metrics of Average Accuracy(Acc), Macro Sensitivity(Se), Macro Specificity(Sp) and Macro Precision(Pre). In Macro method, the average of the accuracy, sensitivity specificity and precision of the system on different subsets are taken, where each subset consists of all images of a specific class. Mathematically they are defined as;
\begin{IEEEeqnarray}{C} \label{eq:acc}
Average \ Accuracy \ =\ \frac{\sum_{i}^{C} \frac{Tp_{i} \ +\ Tn_{i}}{Tp_{i} \ +\ Tn_{i} \ +\ Fp_{i} \ +\ Fn_{i}}}{\sum_{j}^{C} 1}
\end{IEEEeqnarray}

\begin{IEEEeqnarray}{C} \label{eq:se}
Macro - Sensitivity \ =\ \frac{\sum_{i}^{C} \frac{Tp_{i}}{Tp_{i} \ +\ Fn_{i}}}{\sum_{j}^{C} 1}
\end{IEEEeqnarray}

\begin{IEEEeqnarray}{C} \label{eq:sp}
Macro - Specificity \ =\ \frac{\sum_{i}^{C} \frac{Tn_{i}}{Tn_{i} \ +\ Fp_{i}}}{\sum_{j}^{C} 1}
\end{IEEEeqnarray}

\begin{IEEEeqnarray}{C} \label{eq:pre}
Macro - Precision \ =\ \frac{\sum_{i}^{C} \frac{Tp_{i}}{Tp_{i} \ +\ Fp_{i}}}{\sum_{j}^{C} 1}
\end{IEEEeqnarray}

Here, \(Tp_{i}\) represents the true positives; \(Tn_{i}\) represents the true negatives; \(Fp_{i}\) represents the false positives; \(Fn_{i}\) represents the false negatives with respect to the actual and predicted water stress class; such that \(i, j\in Classes\ and\ Classes = \{Before \ Flowering,\ Control,\ Young\ Seedling\}\), and C is the number of classes. 

\section{Experimental Results}
In this section, we describe the four experiments performed on the dataset. Firstly, we examine the water stress classification ability of fine-tuned VGG16 and Inception-V3 by performing time-invariant training and compare it with our previously used technique \cite{azimi2020water}. This experiment also acts as a baseline for temporal analysis, as we use the same CNNs in our CNN-LSTM models. Secondly, we train and evaluate CNN-LSTM models to investigate the effectiveness of temporal analysis of the visual features extracted from plant shoot images. Thirdly, we test the robustness of our model by evaluating it on perturbed sequences of shoot images, such that a certain percentage of images of a sequence undergo Gaussian Noise perturbations. Lastly, we perform an ablation study on the CNN-LSTM model's effectiveness by uniformly decreasing the amount of session data used for training the models. 
\begin{table}[!t]
\caption{Performance metrics for time-invariant water stress classification using CNN models on JG and Pusa varieties of chickpea plants (Acc: Accuracy (in \%), Se: Sensitivity, Sp: Specificity, Pre: Precision).}
\begin{center}
\renewcommand{\arraystretch}{1.2}
\renewcommand{\tabcolsep}{1.6mm}
\begin{tabular}{|c | c | c | c | c | c |}
    \hline
    \hline
    Chickpea Species & CNN Model & Acc & Se & Sp & Pre \\
    \hline
    \multirow{4}{*}{ $JG-62$}&VGG16 & 72.14 & 0.7214 & 0.8734 & 0.7690\\
     \cline{2-6}
      & Inception-V3 & 80.99 & 0.8099 & 0.9135 & 0.8111\\
     \cline{2-6}
      & CNN \cite{azimi2020water} & 78.00 & 0.7800 & 0.8900 & 0.7700\\
    \cline{2-6}
      & ResNet-18 \cite{azimi2020water} & 86.00 & 0.8600 & 0.9300 & 0.8600\\
 \hline
\hline
    \multirow{4}{*}{ $Pusa-372$} & VGG16 & 70.96 & 0.7096 & 0.8737 & 0.7059\\
    \cline{2-6}
    & Inception-V3 & 75.00 & 0.7500 & 0.8750 & 0.7950\\
      \cline{2-6}
     & CNN \cite{azimi2020water} & 76.00 & 0.7600 & 0.8800 & 0.7500\\
     \cline{2-6}
      & ResNet-18 \cite{azimi2020water} & 84.00 & 0.8400 & 0.9200 & 0.8400\\
     \cline{2-6}
\hline
\hline
\end{tabular}
\end{center}
\label{tab:exp1}
\end{table}
\subsection{Time-Invariant Analysis}
In the time-invariant analysis, we train and evaluate four CNN model types, which represent all possible combinations of the plant variety and CNN feature extractor used in this paper, and report the metric scores in Table \ref{tab:exp1}. VGG16 and Inception-V3 fine-tuned networks obtain classification accuracies of $72.14\%$ and $80.99\%$ for JG and $70.96\%$ and $75.00\%$ for Pusa variety, respectively, as shown in Table \ref{tab:exp1}. 

\begin{table}[!t]
\caption{Performance Metrics for temporal water stress classification using CNN-LSTM models on JG and Pusa varieties of chickpea plants using VGG16 and Inception-V3, CNN feature extractor (Acc: Accuracy (in \%), Se: Sensitivity, Sp: Specificity, Pre: Precision).}
\begin{center}
\begin{tabular}{|c | c | c | c | c | c |}
    \hline
    \hline
    Chickpea Species & \makecell{CNN-LSTM \\ Model} & Acc & Se & Sp & Pre \\
    \hline
    \multirow{2}{*}{ $JG-62$} & VGG16 & 98.32 & 0.9833 & 0.9916 & 0.9852\\
      \cline{2-6}
       & Inception-V3 & 98.32 & 0.9833 & 0.9916 & 0.9852\\
     \cline{2-6}
     \hline
\hline
    \multirow{2}{*}{ $Pusa-372$} & VGG16 & 97.50 & 0.9749 & 0.9874 & 0.9778\\
     \cline{2-6}
    & Inception-V3 & 97.50 & 0.9749 & 0.9874 & 0.9778\\
     \hline
\hline
\end{tabular}
\end{center}
\label{tab:exp2}
\end{table}


\subsection{Temporal Analysis}
In the temporal analysis, we train and evaluate four CNN-LSTM model types, which represent all possible combinations of the plant variety and CNN feature extractor used in this paper, and report the metric scores in Table \ref{tab:exp2}. We observe that the classification accuracy of VGG16 and Inception-V3 are $98.32\%$ for JG and $97.5\%$ for Pusa variety. The confusion matrices for each model are shown in Fig. \ref{fig:cf}, where each cell's value represents the average probability across all the folds.

\begin{table}[!t]
\caption{Robustness Analysis of CNN-LSTM models on JG and Pusa varieties of chickpea plants using VGG16 and Inception-V3, CNN feature extractor. Mean(Standard deviation) of the following metrics (Acc: Accuracy (in \%), Se: Sensitivity, Sp: Specificity, Pre: Precision) are reported in this table.}
\begin{center}
\renewcommand{\arraystretch}{1.2}
\renewcommand{\tabcolsep}{1.6mm}
\begin{tabular}{|c | c | c | c | c | c |}
    \hline
    \hline
    Chickpea Species & CNN-LSTM Model & Acc & Se & Sp & Pre \\
    \hline
    \multirow{4}{*}{ $JG-62$} & VGG16 & 
    95.83 & 0.9583 & 0.9789 & 0.9631\\ 
    & & (1.79) & (0.0179) & (0.0091) & (0.0157)\\
     \cline{2-6}
    & Inception-V3 & 95.83 & 0.9583 & 0.9789 & 0.9631\\
    & & (1.79) & (0.0179) & (0.0091) & (0.0157)\\
     \cline{2-6}
    \hline
    \multirow{4}{*}{ $PUSA-372$} & VGG16 & 
    95.16 & 0.9516 & 0.9756 & 0.9572\\ 
    & & (1.57) & (0.0157) & (0.0080) & (0.0138)\\
     \cline{2-6}
    & Inception-V3 & 95.16 & 0.9516 & 0.9756 & 0.9572\\
    & & (1.57) & (0.0157) & (0.0080) & (0.0138)\\
     \cline{2-6}
     \hline
\hline
\end{tabular}
\end{center}
\label{tab:exp3}
\end{table}


\begin{figure}[h!]
\centering
\begin{subfigure}[b]{0.175\textwidth}
\newcommand\items{3} 
\arrayrulecolor{white} 
\noindent\begin{tabular}{cc*{\items}{|E}|}
\multicolumn{1}{c}{} &\multicolumn{1}{c}{} &\multicolumn{\items}{c}{Predicted} \\ \hhline{~*\items{|-}|}
\multicolumn{1}{c}{} & 
\multicolumn{1}{c}{} & 
\multicolumn{1}{c}{\rot{BF}} & 
\multicolumn{1}{c}{\rot{C}} & 
\multicolumn{1}{c}{\rot{YS}} \\ \hhline{~*\items{|-}|}
\multirow{\items}{*}{\rotatebox{90}{Actual}} 
&BF& 1.0 & 0.0& 0.0 \\ \hhline{~*\items{|-}|}
&C& 0.0 & 1.0 & 0.0 \\ \hhline{~*\items{|-}|}
&YS& 0.05 & 0.0 & 0.95 \\ \hhline{~*\items{|-}|}
\end{tabular}
\caption{JG - VGG16}
\end{subfigure}
 \hspace{7.7mm}
 \hfill
\begin{subfigure}[b]{0.165\textwidth}
\newcommand\items{3} 
\arrayrulecolor{white} 
\noindent\begin{tabular}{cc*{\items}{|E}|}
\multicolumn{1}{c}{} &\multicolumn{1}{c}{} &\multicolumn{\items}{c}{Predicted} \\ \hhline{~*\items{|-}|}
\multicolumn{1}{c}{} & 
\multicolumn{1}{c}{} & 
\multicolumn{1}{c}{\rot{BF}} & 
\multicolumn{1}{c}{\rot{C}} & 
\multicolumn{1}{c}{\rot{YS}} \\ \hhline{~*\items{|-}|}
\multirow{\items}{*}{\rotatebox{90}{Actual}} 
&BF& 0.95 & 0.0 & 0.05 \\ \hhline{~*\items{|-}|}
&C& 0.0 & 1.0 & 0.0 \\ \hhline{~*\items{|-}|}
&YS& 0.0 & 0.0 & 1.0 \\ \hhline{~*\items{|-}|}
\end{tabular}
\caption{JG - Inception-V3}
\end{subfigure}
\qquad \qquad \quad
\par\bigskip
 \bigskip
 \begin{subfigure}[b]{0.175\textwidth}
\newcommand\items{3}   
\arrayrulecolor{white} 
\noindent\begin{tabular}{cc*{\items}{|E}|}
\multicolumn{1}{c}{} &\multicolumn{1}{c}{} &\multicolumn{\items}{c}{Predicted} \\ \hhline{~*\items{|-}|}
\multicolumn{1}{c}{} & 
\multicolumn{1}{c}{} & 
\multicolumn{1}{c}{\rot{BF}} & 
\multicolumn{1}{c}{\rot{C}} & 
\multicolumn{1}{c}{\rot{YS}} \\ \hhline{~*\items{|-}|}
\multirow{\items}{*}{\rotatebox{90}{Actual}} 
&BF& 1.0   & 0.0  & 0.0   \\ \hhline{~*\items{|-}|}
&C & 0.0   & 1.0  & 0.0   \\ \hhline{~*\items{|-}|}
&YS & 0.0   & 0.075   & 0.925   \\ \hhline{~*\items{|-}|}
\end{tabular}
\caption{Pusa - VGG16}
\end{subfigure}
 \hspace{7.7mm}
 \hfill
\begin{subfigure}[b]{0.165\textwidth}
\newcommand\items{3} 
\arrayrulecolor{white} 
\noindent\begin{tabular}{cc*{\items}{|E}|}
\multicolumn{1}{c}{} &\multicolumn{1}{c}{} &\multicolumn{\items}{c}{Predicted} \\ \hhline{~*\items{|-}|}
\multicolumn{1}{c}{} & 
\multicolumn{1}{c}{} & 
\multicolumn{1}{c}{\rot{BF}} & 
\multicolumn{1}{c}{\rot{C}} & 
\multicolumn{1}{c}{\rot{YS}} \\ \hhline{~*\items{|-}|}
\multirow{\items}{*}{\rotatebox{90}{Actual}} 
&BF& 0.925 & 0.075 & 0.0 \\ \hhline{~*\items{|-}|}
&C& 0.0 & 1.0 & 0.0 \\ \hhline{~*\items{|-}|}
&YS& 0.0 & 0.0 & 1.0 \\ \hhline{~*\items{|-}|}
\end{tabular}
\caption{Pusa - Inception-V3}
\end{subfigure} 
\qquad \qquad \quad
\caption{Confusion matrix depicting the results for CNN-LSTM model with VGG16 and Inception-V3 as the CNN feature extractor trained on 32 sessions of JG and Pusa species. Here, (a), (b), (c), and (d) represent the confusion matrices of the different possible species - feature extractor models.}
\label{fig:cf}
\end{figure}

\begin{table*}[!t]
\caption{Ablation Study: Performance metrics for CNN-LSTM model on JG-62 and Pusa-372 varieties using VGG16 and Inception-V3 image feature extractor (Acc: Accuracy (in \%), Se: Macro-Sensitivity, Sp: Macro-Specificity, Pre: Macro-Precision) on \emph{Sn} session data where n represent images of dataset up to the nth session.}
\begin{center}
\arrayrulecolor{black} 
\renewcommand{\arraystretch}{1.2}
\begin{tabular}{|c | c | c | c | c | c |c | c | c | c | c |}
    \hline
    \hline
    Chickpea Species & Feature Extractor & Metric & S4 & S8 & S12 & S16 & S20 & S24 & S28 & S32\\
    \hline
   \multirow{8}{*}{ $JG-62$}& \multirow{4}{*}{ VGG16} & Acc & 87.5 & 91.67 & 93.33 & 93.33 & 95.83 & 97.5 & 97.5 & 98.32 \\
     \cline{3-11}
       & & Se & 0.875& 0.9167 & 0.9333 & 0.9333 & 0.9583	& 0.9749 & 0.9749 & 0.9833 \\
     \cline{3-11}
       && Sp & 0.9375 & 0.9583	& 0.9662 & 0.9662 & 0.9791 & 0.9874	& 0.9874 & 0.9916 \\
     \cline{3-11}
    & & Pre & 0.8857 & 0.9267 & 0.9412 & 0.9557	& 0.963	& 0.9704 & 0.9778& 0.9852 \\
     \cline{2-11}
     &\multirow{4}{*}{Inception-V3} & Acc & 87.5 & 91.67 & 93.33 & 93.33  & 95.83 & 97.5 & 97.5 & 98.32 \\
     \cline{3-11}
        && Se & 0.875& 0.9167 & 0.9333 & 0.9333 & 0.9583	& 0.9749 & 0.9749 & 0.9833 \\
     \cline{3-11}
      && Sp & 0.9375 & 0.9583	& 0.9662 & 0.9662 & 0.9791 & 0.9874	& 0.9874 & 0.9916 \\
     \cline{3-11}
    && Pre &  0.8857 & 0.9267 & 0.9412 & 0.9557	& 0.963	& 0.9704 & 0.9778& 0.9852 \\
\hline
\hline
  \multirow{8}{*}{ $Pusa-372$}& \multirow{4}{*}{ VGG16}& Acc & 83.33	& 87.5	& 91.67	& 93.33	& 95.83	& 96.66	& 96.66 & 97.5 \\
      \cline{3-11}
       & & Se & 0.8333 & 0.875 & 0.9167 & 0.9333 & 0.9583	& 0.9666 & 0.9666 & 0.9749 \\
     \cline{3-11}
       & & Se & 0.9167	& 0.9583 & 0.9583 & 0.9666 & 0.9791	& 0.9833 & 0.9833 & 0.9874 \\
     \cline{3-11}
    & & Pre & 0.8426 & 0.933 & 0.933 & 0.945 & 0.963 & 0.9704 & 0.9704 & 0.9778 \\
     \cline{2-11}
     &\multirow{4}{*}{Inception-V3}  & Acc & 87.5 & 93.33 & 94.99 & 95.83 & 96.66 & 96.66 & 97.5 & 97.5 \\
     \cline{3-11}
        & & Se & 0.8751 & 0.9083 & 0.9249 & 0.9583	& 0.9666 & 0.9666 & 0.9749 & 0.9749 \\

     \cline{3-11}
     & & Sp & 0.9375	& 0.9666 & 0.9707 & 0.9791 & 0.9833	& 0.9833 & 0.9874 & 0.9874 \\

     \cline{3-11}
     & & Pre & 0.8857 & 0.945 & 0.951 & 0.963 & 0.9704	& 0.9704 & 0.9778 & 0.9778 \\
\hline
\hline
\end{tabular}
\end{center}
\label{tab:ablation}
\end{table*}

\begin{figure*}[!t]
\Centering
\includegraphics[width=13cm, height=13cm]{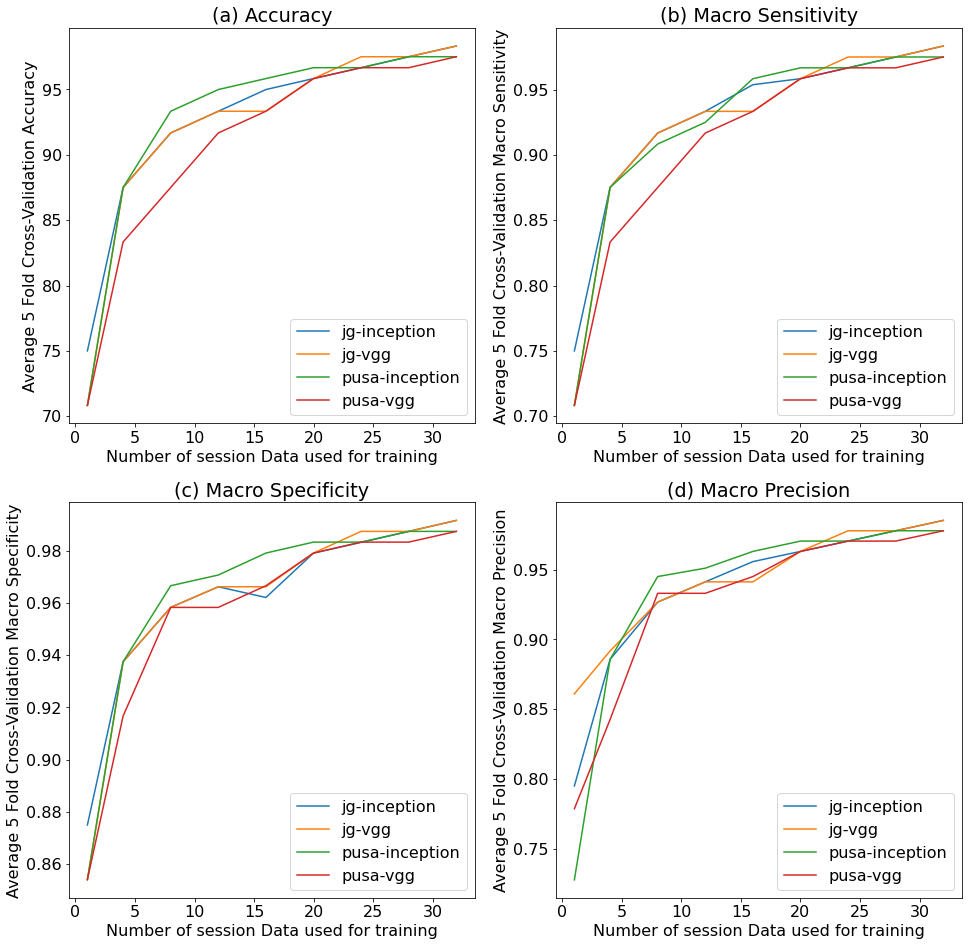}
\caption{Visualizing the accuracy, macro-sensitivity, macro-specificity, and macro-precision of models trained on different chickpea plant species - feature vector combination over the number of sessions data for training. Here (a), (b), (c), and (d) represent accuracy, macro-sensitivity, macro-specificity, and macro-precision, respectively.}
\label{fig:all_graphs}
\end{figure*}

\subsection{Robustness Analysis}
\label{sec:ranalysis}
In this experiment, we add Gaussian noise perturbations to the test sequences of each cross-validation fold. We start by selecting a certain percentage of images from each test sequence in random order and perturb them with noise. We often come across a range of image perturbation percentages rather than a fixed value in real-life scenarios. The minimum percentage is nearly 3\% or 1 out of 32 images in a sequence. The maximum perturbation percentage is set to nearly one-third of images of an entire test sequence, approximately equal to 10 out of 32 images. For perturbations greater than one-third of the images in a test sequence, applying a noise removal preprocessing step before training the CNN-LSTM model will be computationally more efficient than training a larger and more complex neural model, which is inherently unaffected by noise. After selecting the images, we apply noise to them. The noise intensities are sampled from the distribution described in section \ref{eq:gauss}. Then, we perform ten model evaluation cycles by increasing the number of perturbed images in a test sequence from 1 to 10, with an increment of one image per cycle. Finally, we report the mean accuracy and standard deviation across all cycles as shown in Table \ref{tab:exp3}.


\subsection{Ablation Study}
\label{sec:abstudy}
We perform an ablation study to determine the CNN-LSTM model's performance on decreasing the session data used for training. In this study, we evaluate CNN-LSTM models corresponding to each chickpea plant species \({JG, Pusa}\), and CNN feature extractor \({VGG16, Inception-V3}\) pair. We train 8 models for each pair, such that each model differs in the number of session data used. Starting from the 32nd session down to the 4th session, we reduce the number of sessions data by 4. A gap of 4 sessions was chosen as it provided the best solution for the trade-off between the available computational resources and time for training models vs. the change in the performance metrics' value between two consecutive models. We report the results obtained in Table \ref{tab:ablation} and visualize the value of each performance metric vs. the number of session data in the graphs shown in Fig. \ref{fig:all_graphs}.

\subsection{Computational Complexity}
In this section, we report the time and space complexity of inference. To report the worst-case complexities, we utilize the models trained on 32 sessions of data, as these models have the maximum number of parameters. The inference time does not include the time to load the 32 session images, pre-process the image, and load the model. In other words, we measure the time taken for the feedforward propagation of the model. We calculate the inference time on the Nvidia Tesla K80 GPU and Intel(R) Xeon(R) CPU. Our CNN-LSTM models with VGG16 feature extractor and Inception-V3 feature extractors have nearly 17M  and 56M parameters, respectively. Further, the model with the VGG16 feature extractor takes \textbf{29ms} and \textbf{70ms} to predict the plant's water stress condition on the GPU and CPU, respectively. Whereas the model with the Inception-V3 feature extractor, which has more parameters, takes \textbf{59ms} and \textbf{98ms} to predict the plant's water stress condition on the GPU and CPU, respectively. 

\section{Discussion}
This section presents the discussion on the experimental results, application scope of this research, and its limitations.

\subsection{Experimental Inference}
This subsection provides a discussion on the results of the four experiments presented in this paper.

\subsubsection{Time Invariant Analysis}
Firstly, we observe that VGG16 and Inception-V3 models' performance is similar to the performance of our previous techniques, that is, a custom-CNN architecture and ResNet-18 architecture (as shown in Table \ref{tab:exp1}). ResNet-18, with fewer parameters than Inception-V3 and VGG16 models, has better water stress classification performance due to a higher degree of overfitting in the latter two larger models. Secondly, we infer that VGG16 and Inception-V3 models' performance on JG images is better than Pusa images, which is consistent with our previous results, as shown in Table \ref{tab:exp1}. This can be attributed to the water-resistant nature of Pusa species and the water-sensitive nature of JG species. In other words, the visual changes introduced due to water stress are more prominent in JG than Pusa, thus making it easier to classify JG images into the three water stress categories. Lastly, we observe that Inception-V3 models produced better results than VGG16 models across both chickpea species. This can be explained by relating these results with both these architectures' classification results on the ImageNet dataset. Inception-V3 outperforms VGG16 in both the top 1 and top 5 error rates (\%) because it can extract better visual features \cite{inceptionv3, vgg16}. This suggests that image-based classification by transfer learning from Inception-V3 should be better than VGG16, consistent with the results shown in Table \ref{tab:exp1}. Therefore, time-invariant water-stress classification is network-dependent.

\begin{figure*}[!h]
	\begin{minipage}{.5\textwidth}
	    \centering  
        \begin{subfigure}[t]{.13\textwidth}
            \centering
    		\includegraphics[width=1.5cm,height=1.75cm]{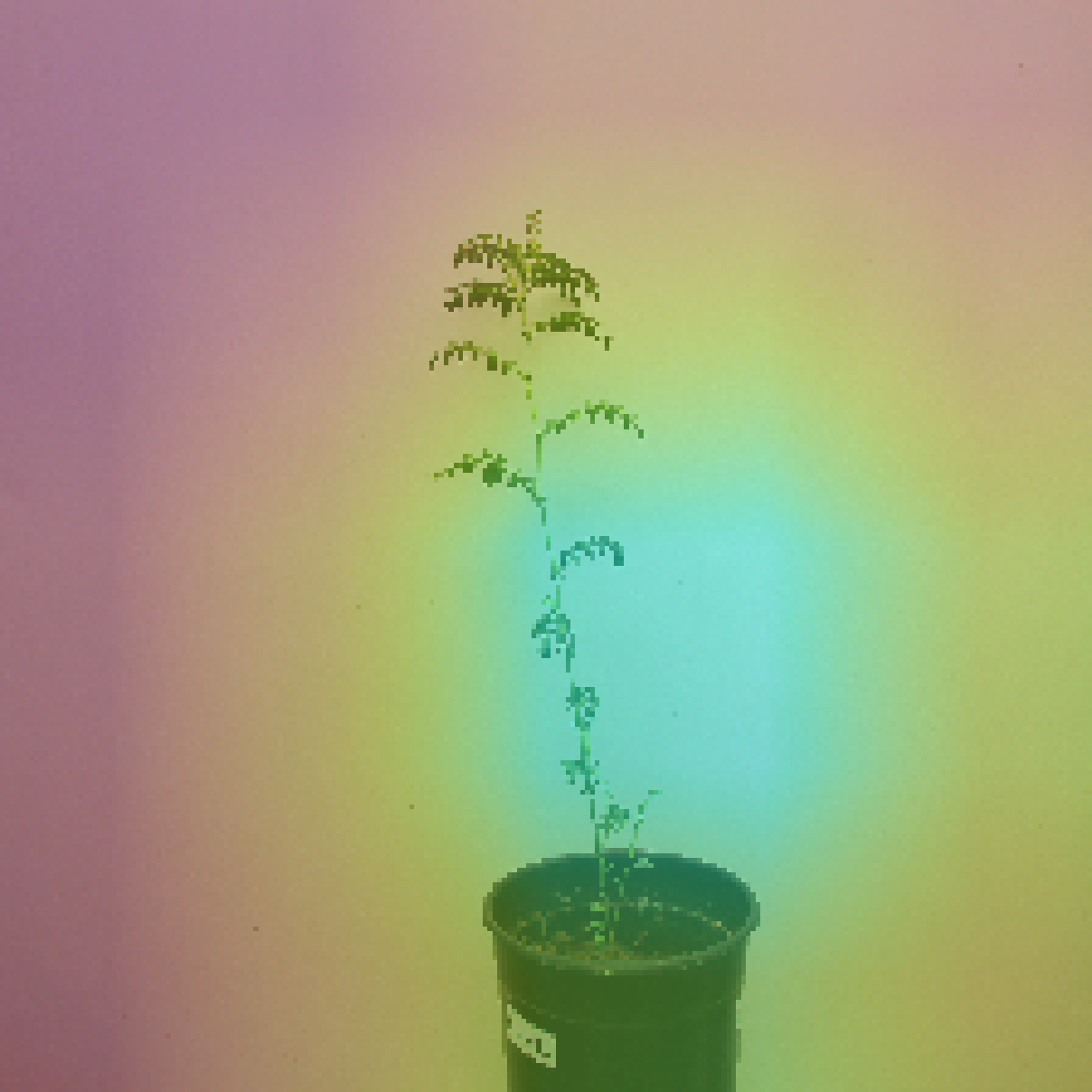}
    		\caption{}
	    \end{subfigure}
	    \hspace{0.15cm}
	    \begin{subfigure}[t]{.13\textwidth}
            \centering
            \includegraphics[width=1.5cm,height=1.75cm]{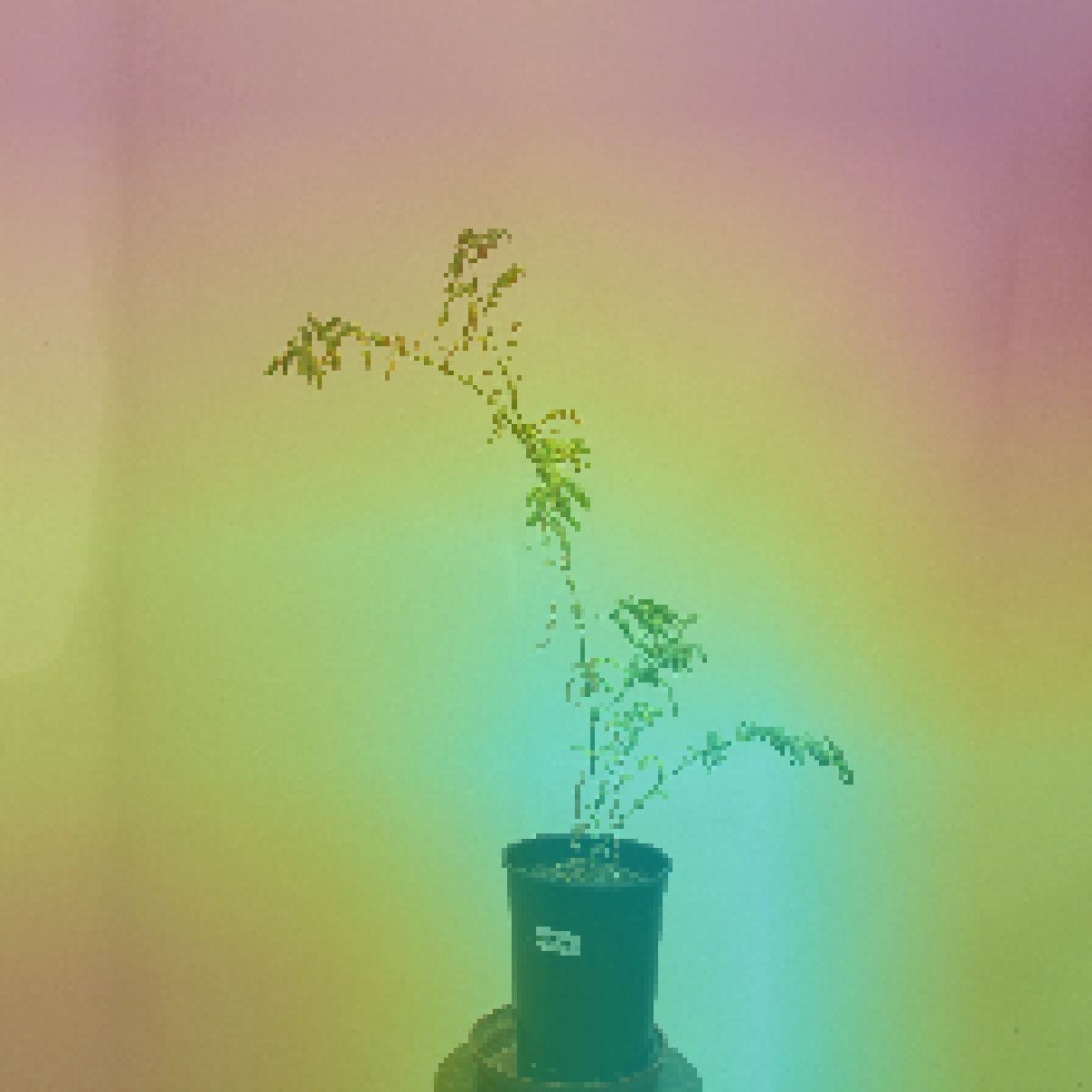}
        	\caption{}
	    \end{subfigure}
	    \hspace{0.15cm}
	    \begin{subfigure}[t]{.13\textwidth}
        \centering
            \includegraphics[width=1.5cm,height=1.75cm]{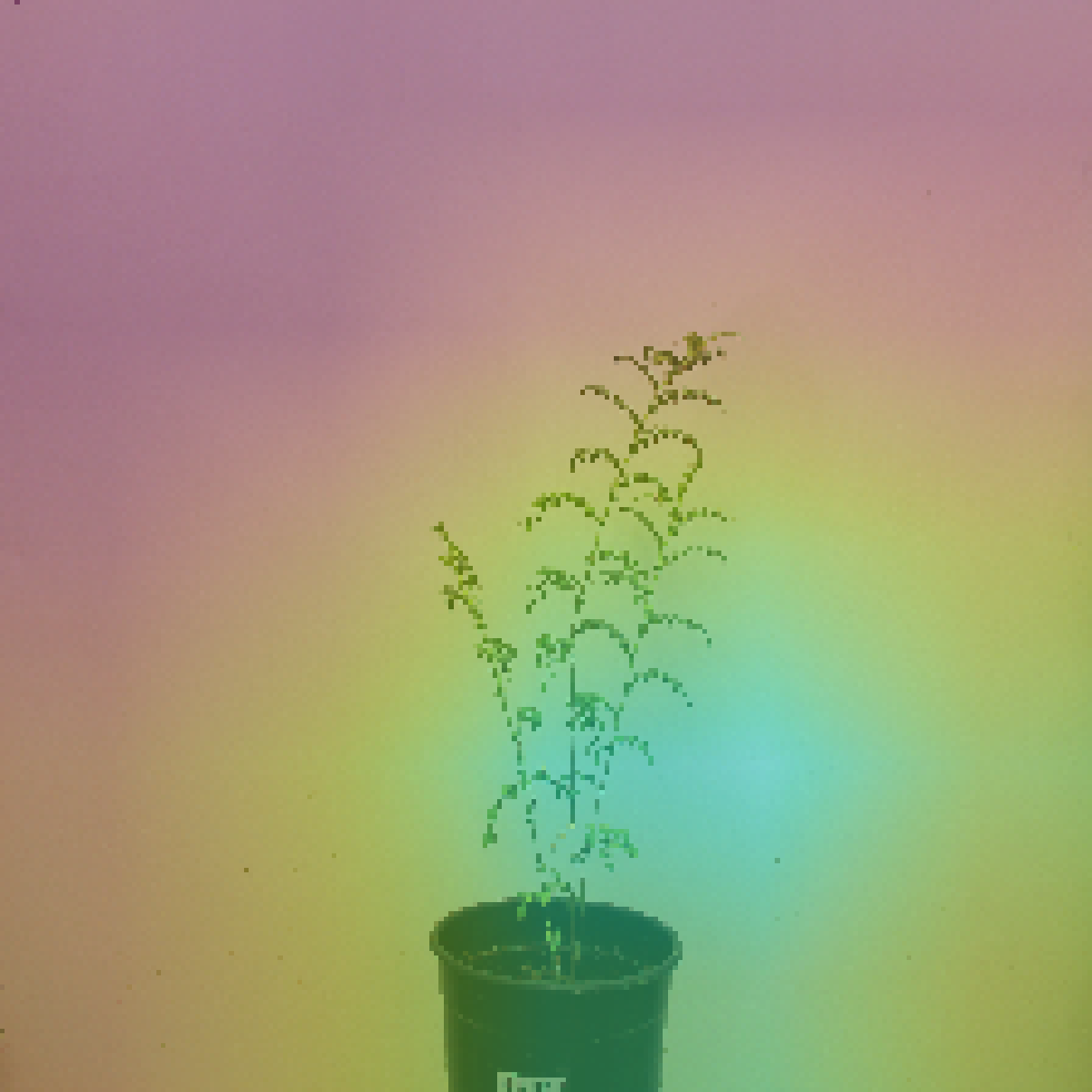}
        	\caption{}
	    \end{subfigure}%
	    \hspace{0.25cm}
	    \begin{subfigure}[t]{.13\textwidth}
            \centering
            \includegraphics[width=1.5cm,height=1.75cm]{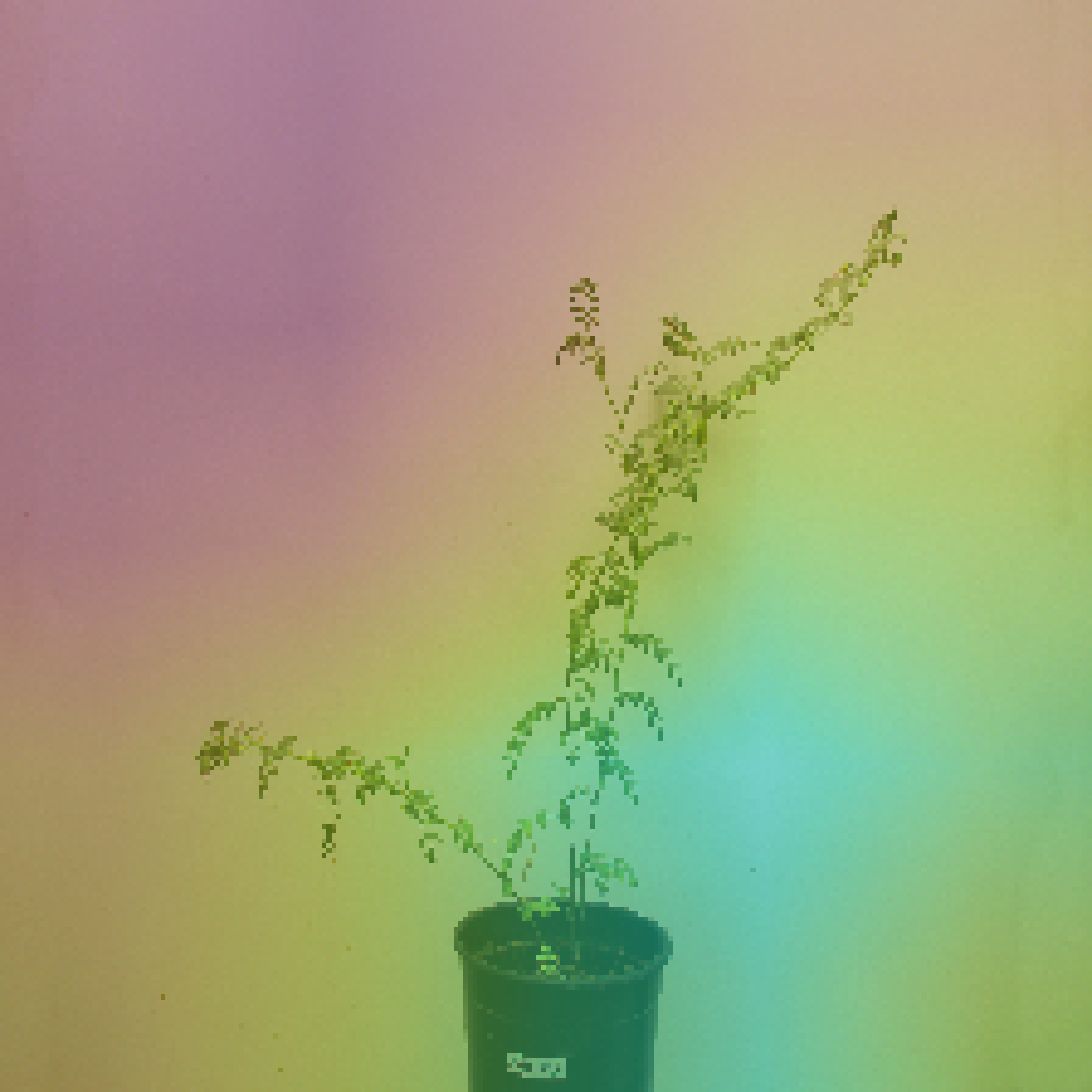}
        	\caption{}
    	\end{subfigure}%
	    \hspace{0.25cm}
	    \begin{subfigure}[t]{.13\textwidth}
            \centering
            \includegraphics[width=1.5cm,height=1.75cm]{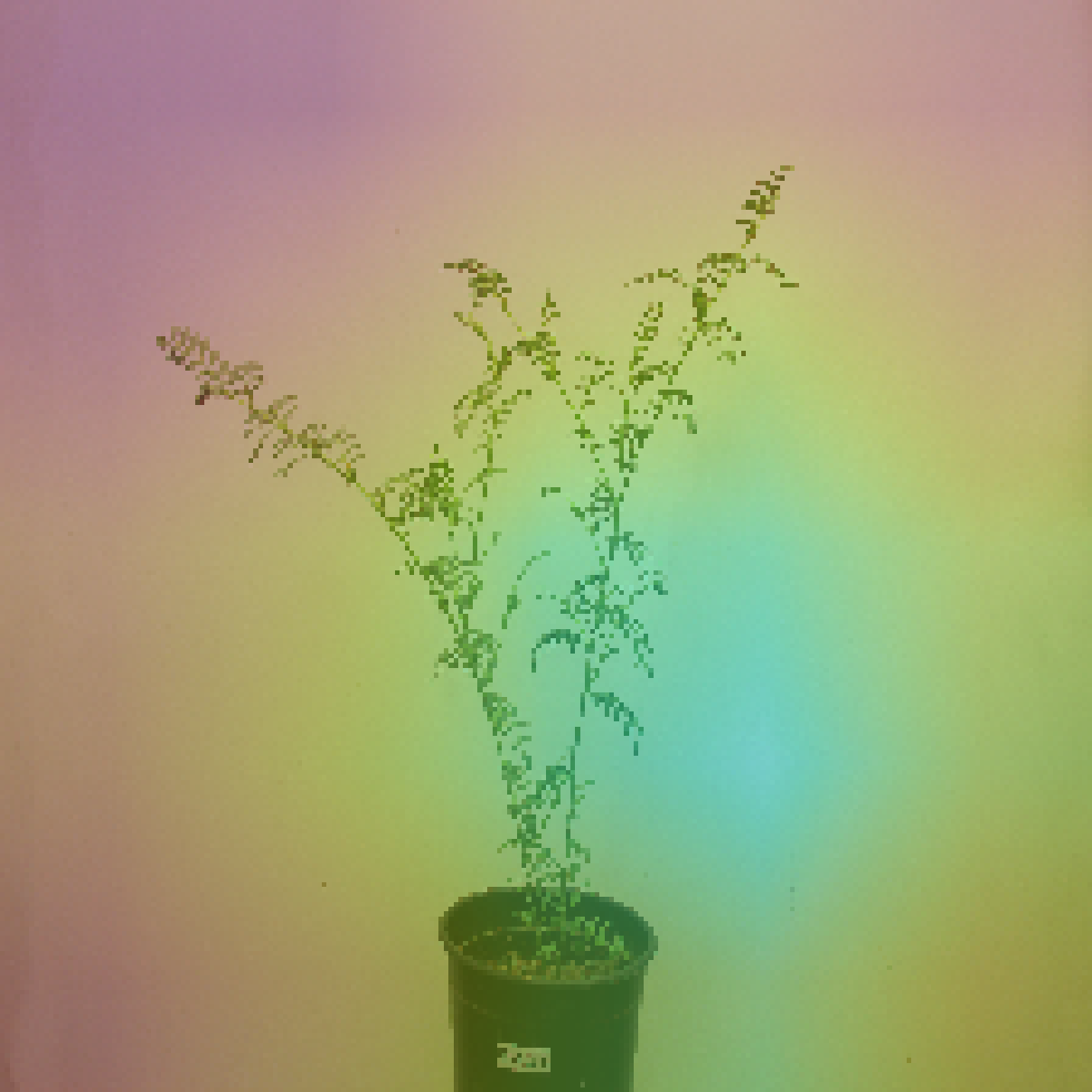}
        	\caption{}
    	\end{subfigure}%
	    \hspace{0.25cm}
	    \begin{subfigure}[t]{.13\textwidth}
            \centering
            \includegraphics[width=1.5cm,height=1.75cm]{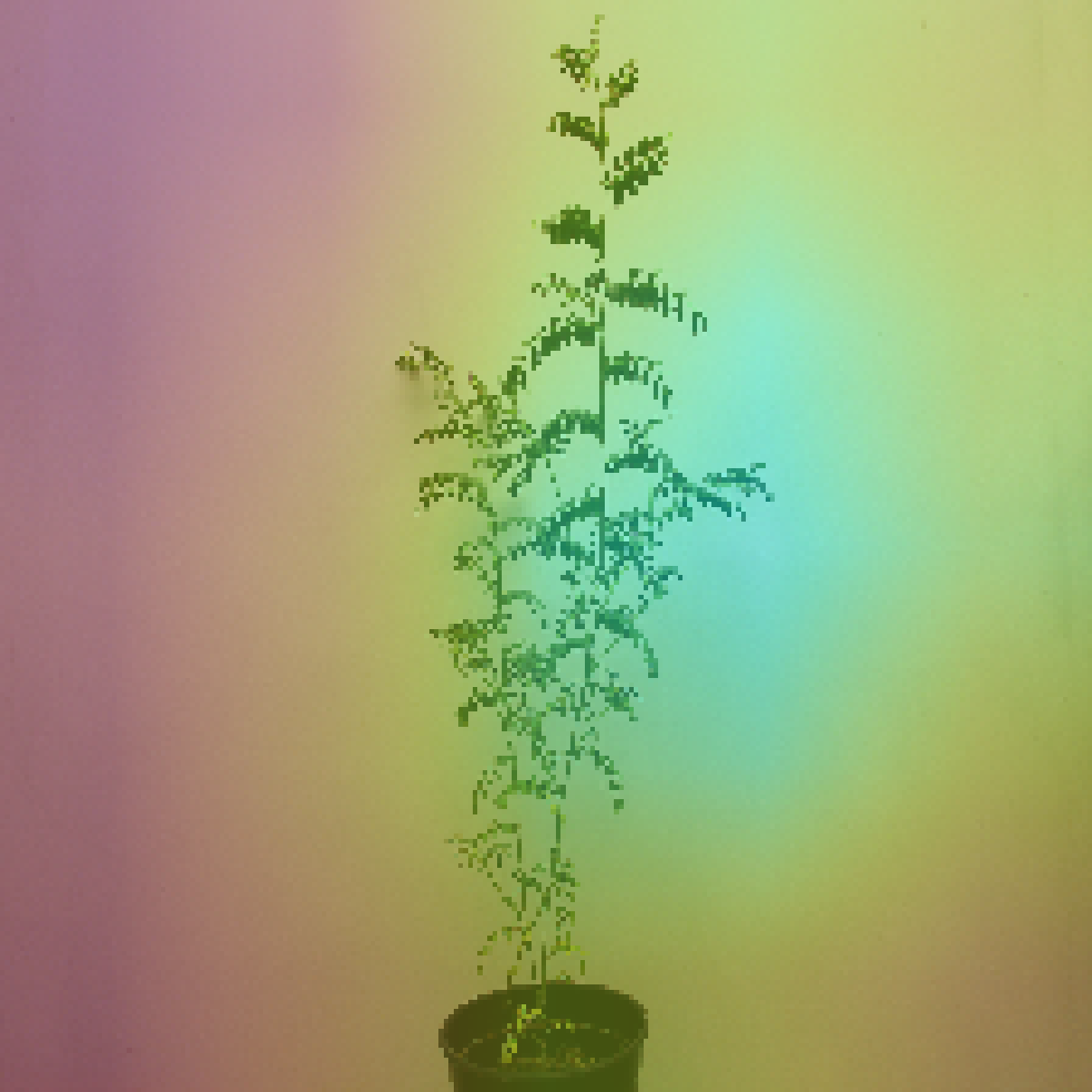}
    		\caption{}
    	\end{subfigure}%
    	\caption{Grad-CAM visualization of JG-62 images, with respect to Inception V3 CNN feature extractor. Figures (a), (b) belong to Young Seedling; (c), (d) belong to Before Flowering; (e), (f) belong to Control.}
    	\label{fig:jg_gradcam}
	\end{minipage}
	\hspace{0.5cm}
	\begin{minipage}{.5\textwidth}
		\centering  
		 \begin{subfigure}[t]{.13\textwidth}
            \centering
    		\includegraphics[width=1.5cm,height=1.75cm]{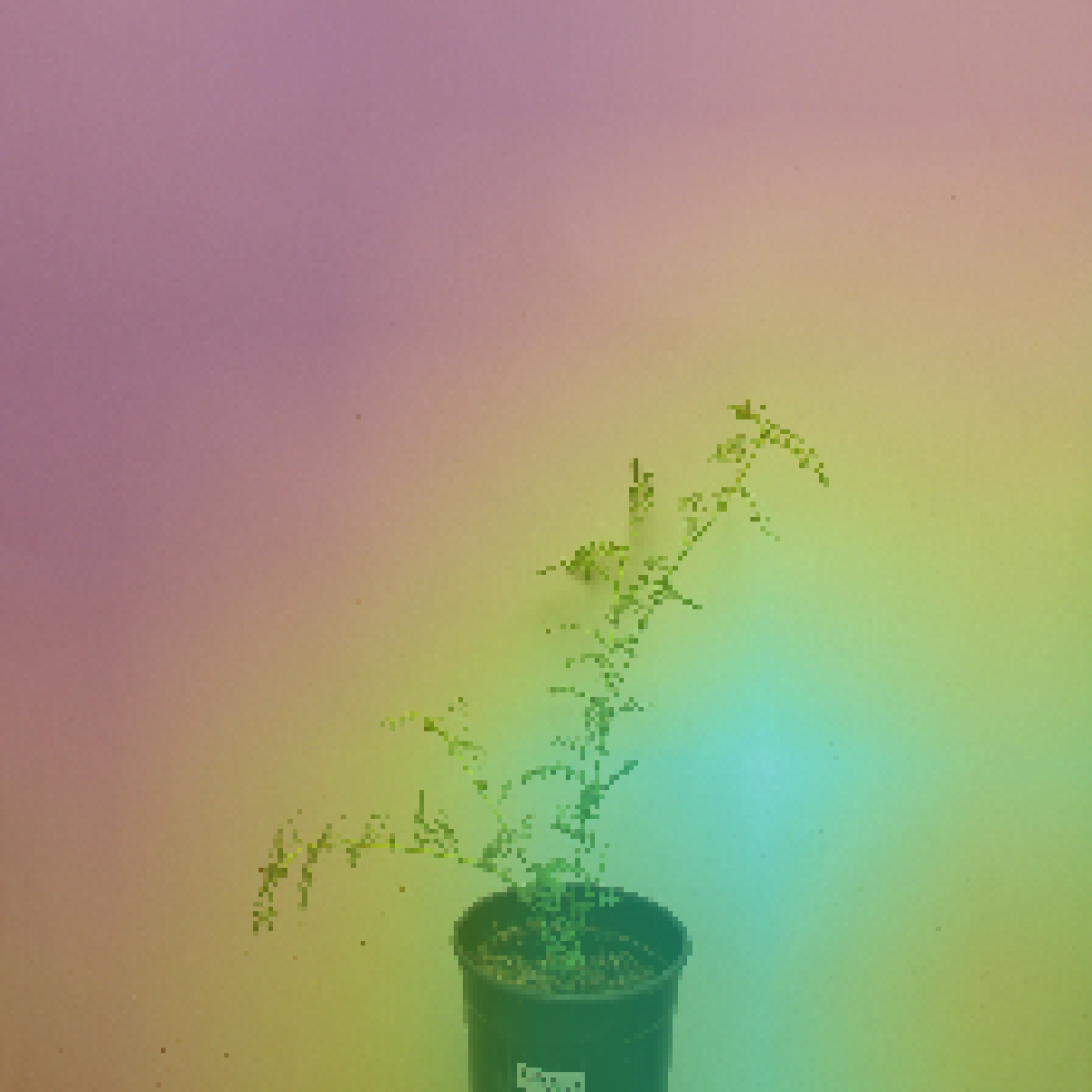}
    	    \caption{}
	    \end{subfigure}
	    \hspace{0.15cm}
	    \begin{subfigure}[t]{.13\textwidth}
            \centering
            \includegraphics[width=1.5cm,height=1.75cm]{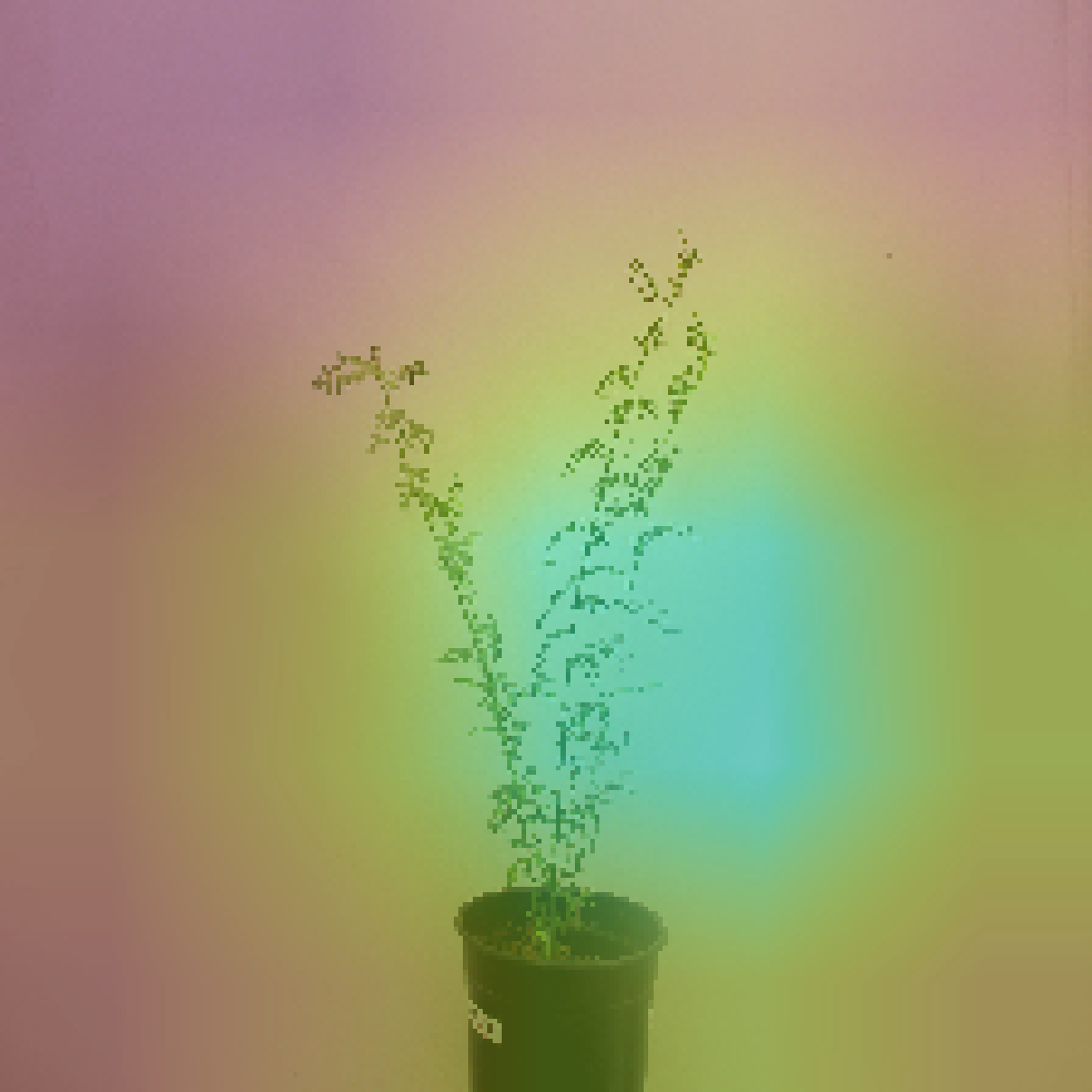}
            \caption{}
	    \end{subfigure}
	    \hspace{0.15cm}
	    \begin{subfigure}[t]{.13\textwidth}
        \centering
            \includegraphics[width=1.5cm,height=1.75cm]{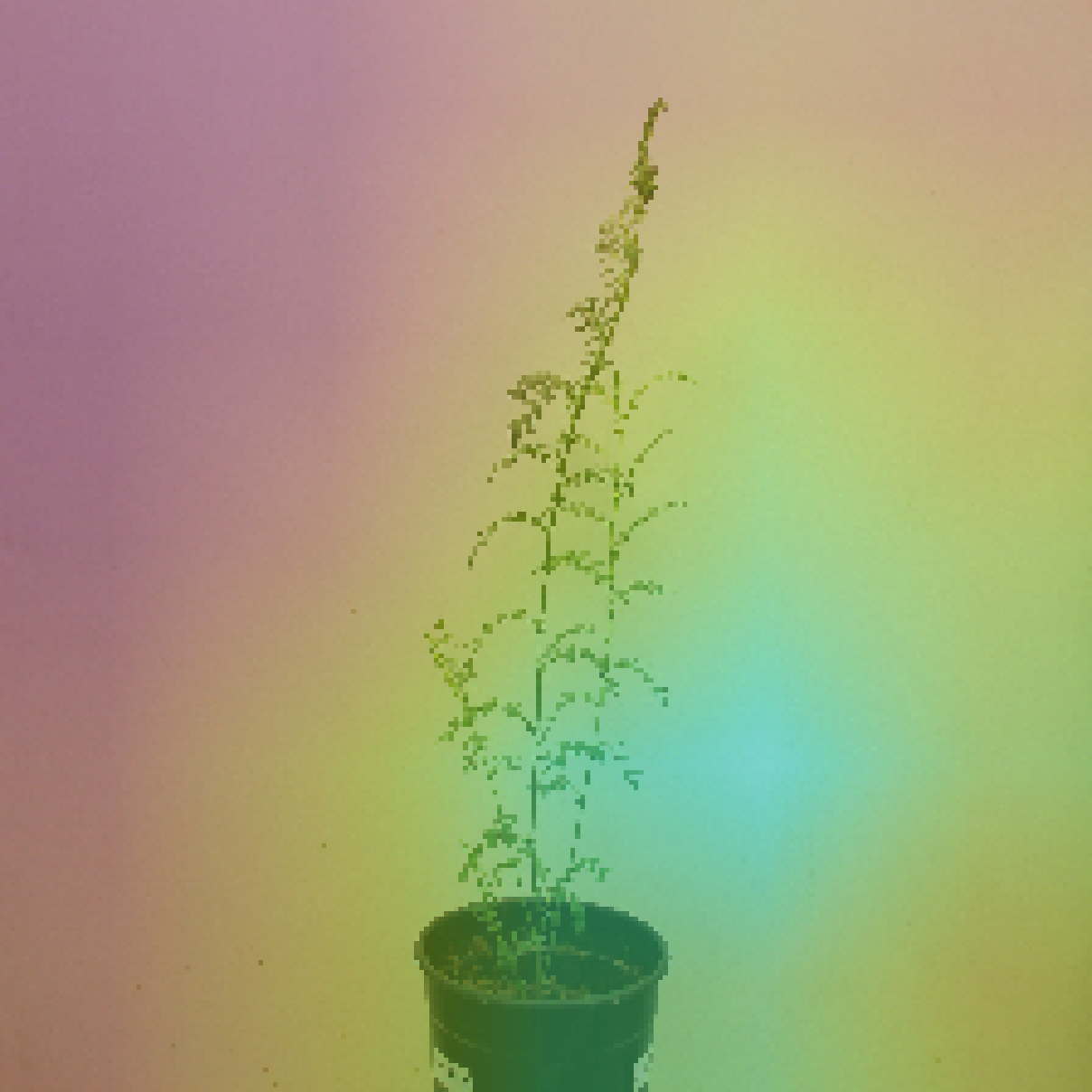}
    		\caption{}
	    \end{subfigure}%
	    \hspace{0.25cm}
	    \begin{subfigure}[t]{.13\textwidth}
            \centering
            \includegraphics[width=1.5cm,height=1.75cm]{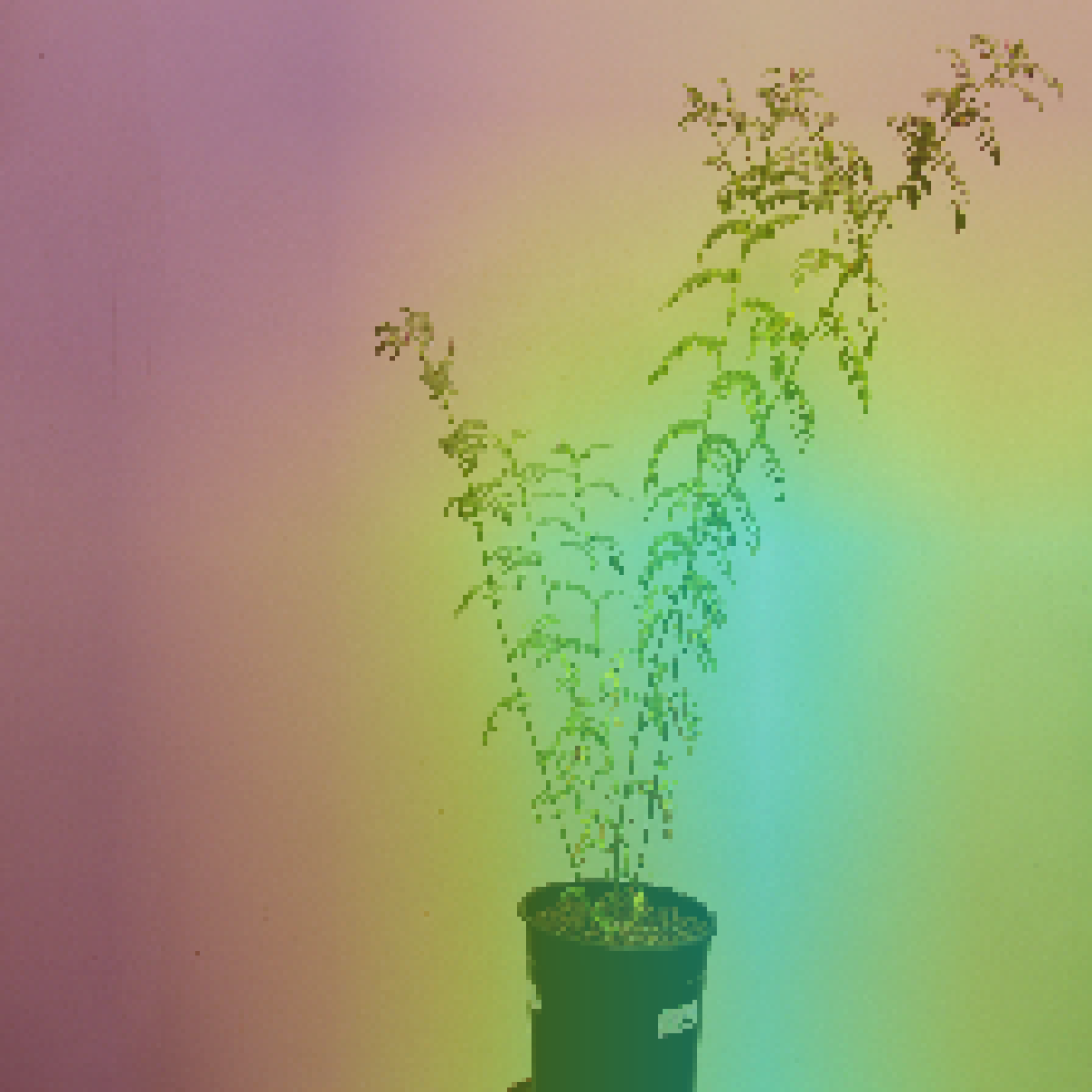}
    		\caption{}
    	\end{subfigure}%
	    \hspace{0.25cm}
	    \begin{subfigure}[t]{.135\textwidth}
            \centering
            \includegraphics[width=1.5cm,height=1.75cm]{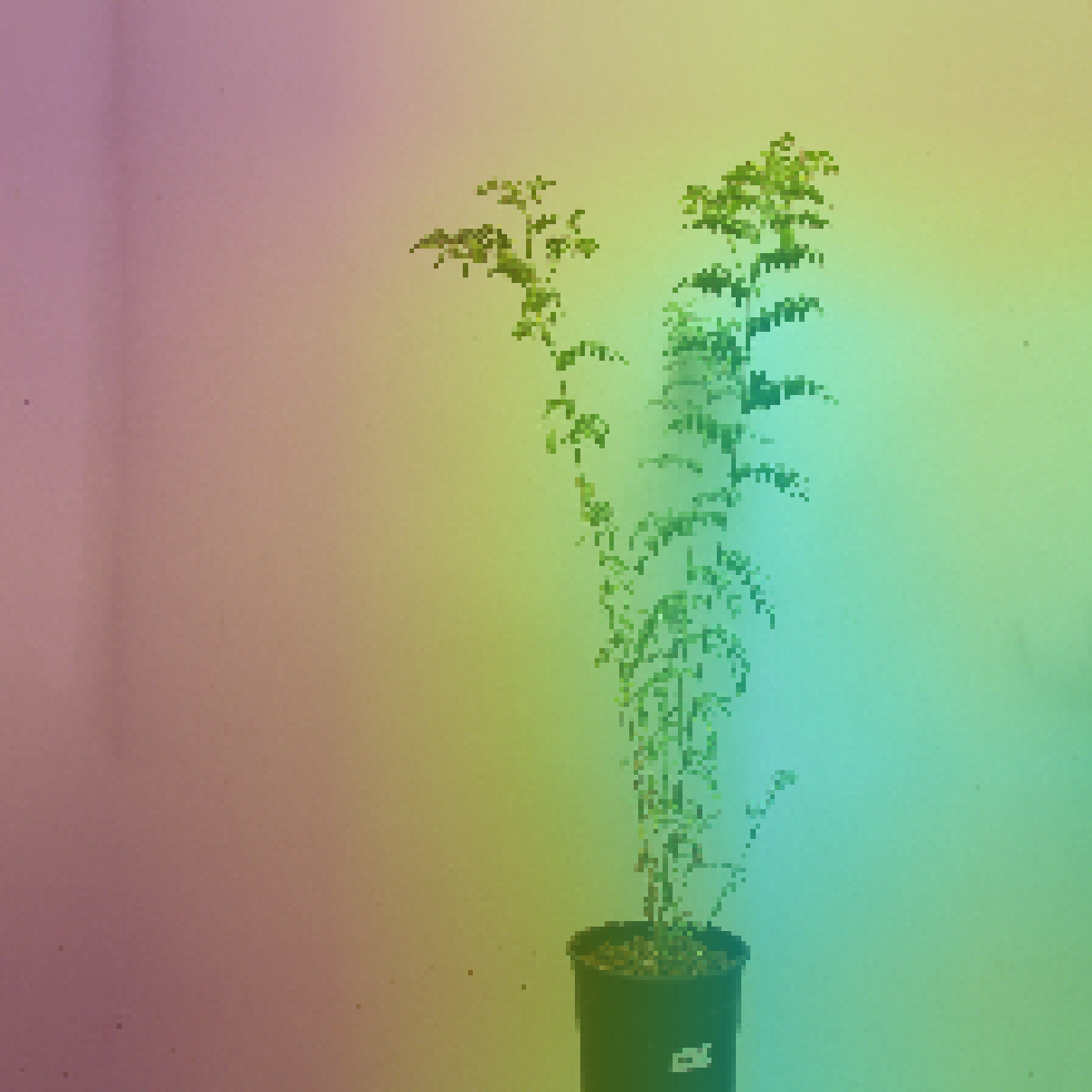}
    		\caption{}
    	\end{subfigure}%
	    \hspace{0.25cm}
	    \begin{subfigure}[t]{.13\textwidth}
            \centering
            \includegraphics[width=1.5cm,height=1.75cm]{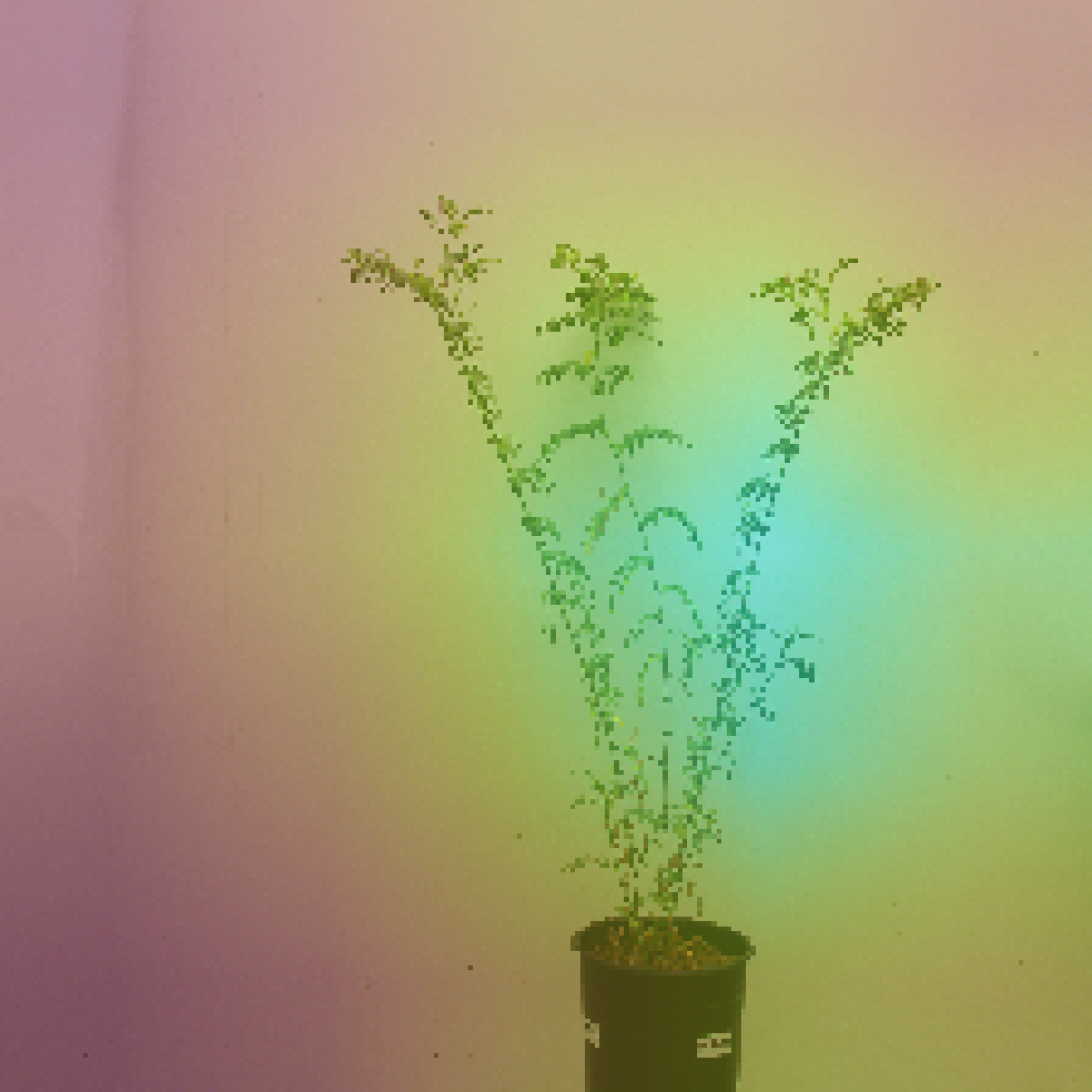}
    		\caption{}
    	\end{subfigure}%
    	\caption{Grad-CAM visualization of Pusa-372 images, with respect to Inception V3 CNN feature extractor. Figures (a), (b) belong to Young Seedling; (c), (d) belong to Before Flowering; (e), (f) belong to Control.}
    	\label{fig:pusa_gradcam}
	\end{minipage}
\end{figure*}

\subsubsection{Temporal Analysis}
On comparing the results of temporal analysis of visual changes induced in the chickpea plant shoots due to water stress (shown in Table \ref{tab:exp2}) with the time-invariant analysis (shown in Table \ref{tab:exp1}), we observe that temporal analysis outperforms the best reported time-invariant scores by at least 14\% for both chickpea varieties. Additionally, each chickpea variety's results are consistent for both feature extractors, showing that the proposed CNN-LSTM technique is independent of the feature extractor network. We also observe that the CNN-LSTM performs better on JG than Pusa. This observation corresponds to the inherent water stress sensitivity characteristics of these two chickpea varieties. JG is water stress-sensitive, thus producing more noticeable visual changes over time than Pusa, which is water stress-tolerant. This can also be seen from the time-invariant analysis results in this paper, as shown in Table \ref{tab:exp1}. 

\subsubsection{Feasability analysis of CNN feature extractor using Grad-CAM}

This section examines the feasibility of the CNN feature extractor considered in this paper. We employ Gradient-weighted Class Activation Mapping (Grad-CAM) \cite{vis-gradcam}. Grad-CAM uses the gradient information flowing into the last convolutional layer of CNN to understand each neuron for a class label of interest. As we use the same CNN feature extractors for time-invariant and our proposed temporal analysis, Grad-CAM visualization of the CNN network used for time-invariant analysis can be used to approximately extrapolate the behavior of these extractors in the temporal context. Further, we use the CNN network with Inception-V3 feature extractor, the best-reported network by this work. While applying Grad-CAM, we obtain the class discriminative localization map of width u and height v for a water stress class c by first computing the gradient of the score for that class, that is, \({y}^{c}\) (before the softmax), for feature maps \(\alpha_{k}\) of a convolutional layer. These gradients flowing back are global average-pooled over the width and height dimensions (indexed by i and j respectively) to obtain the neuron importance weights \(\alpha_{k}^{c}\). 

\begin{equation}\label{eq:gradcam1}
\alpha_{k}^{c}=\overbrace{\frac{1}{Z} \sum_{i} \sum_{j}}^{\text {global\ average\ pooling }} \underbrace{\frac{\partial y^{c}}{\partial A_{i j}^{k}}}_{\text {gradients\ via\ backprop }}
\end{equation}

After calculating \(\alpha_{k}^{c}\), we perform a weighted combination of the activation maps and follow it by a ReLU. Without it, the class activation map highlights more than required and achieves low localization performance. 

\begin{equation}\label{eq:gradcam2}
L_{\mathrm{Grad}-\mathrm{CAM}}^{c}=\operatorname{ReLU} \underbrace{\left(\sum_{k} \alpha_{k}^{c} A^{k}\right)}_{\text {linear \ combination }}
\end{equation}

Subsequently, we superimpose the activation map (heatmap) with the original image to coarsely visualize which region it focuses on to classify water stress. In the images shown in Fig. \ref{fig:jg_gradcam} and \ref{fig:pusa_gradcam}, the intensity of yellow is directly proportional to the intensity of neural activation with respect to the predicted class. In other words, CNN focuses on the yellow highlighted regions in the image to make its prediction. From the figures, we observe that the CNN focuses on the shoot of the chickpea plant to predict water stress for both varieties. Further, this area of focus varies with the size and shape of the shoot. These visualizations explain and validate the use of chickpea plant shoot images to detect water stress.

\subsubsection{Robustness Analysis}
On comparing the results in Table \ref{tab:exp2} and Table \ref{tab:exp3}, we observe that the mean accuracy of our model on noisy test data is less than the accuracy on noise-free test data, by atmost 2.5\%. This decrease is consistent for JG and Pusa varieties and VGG16 and Inception-V3 feature extractors.  Even in the presence of noise, each chickpea variety's results are consistent for both the feature extractors, thereby highlighting that the temporal technique is independent of the feature extractor. The model accuracy on the JG variety is greater than that of Pusa, which further validates the water-sensitive nature of JG over the Pusa variety. An interesting observation is the small standard deviation from the mean accuracy for all the CNN-LSTM models. A small standard deviation demonstrates that the model will not be adversely affected by noise, and its accuracy will remain reasonably consistent. Thus, the small decrease in classification accuracy and a fairly consistent average accuracy in noisy conditions makes this technique suitable for real-time deployment.

\subsubsection{Ablation Study}
We draw the following inferences from the result. Firstly, the graphs in Fig. \ref{fig:all_graphs} and Table \ref{tab:ablation} demonstrate that by decreasing the number of session data for training the model, we decrease its ability to discern water stress conditions. This observation is reasonable because a longer image sequence will learn better differentiating features, especially since water stress on the shoot is prominent in the later stages of growth. Secondly, the performance metric curves (as shown in Fig \ref{fig:all_graphs}) for a given plant species are similar for both the feature extractors. This emphasizes that temporal analysis using CNN-LSTM models has negligible dependence on the CNN feature extractor used. On the contrary, the time-invariant classification of water stress depends on the CNN architectures employed, as shown in Table \ref{tab:exp1}. This observation further reinforces the merit of temporal analysis. Lastly, we also observe that these curves and final scores are similar across both species, thereby demonstrating that temporal analysis performs well across different chickpea plant species. Species invariance is another beneficial characteristic for the real-time deployment of this technique.

\subsection{Application}
The image and deep learning-based water stress classification methods can detect the lack of water in plants and its excess. This can help farmers optimize irrigation, which will, in turn, prevent unnecessary expenditure and promote optimum productivity by ensuring good soil health. 
Our deep learning pipeline focuses on solving water stress due to water deficiency crops may face during the growth period. In our dataset, we use a single chickpea plant per image and fluorescent lighting to simulate daylight conditions. Then, we train an CNN-LSTM model to learn visual spatial-temporal features that help classify water stress. To use our approach in real-time, we will require images of an individual crop from a field, taken over time. This will act as real-time test data. We can repeat this process for multiple crops in the field to get a general idea about the water stress situation. 

\subsection{Limitations}
Our proposed deep learning pipeline has shown merits in the form of high water stress identification performance, robustness to noisy conditions, and independence from the type of visual feature extractor used. However, it does have a couple of limitations. Firstly, we train the CNN-LSTM model on our dataset that simulates daylight conditions during photo capturing sessions. Thus, the model may show a slight variation in performance when actually used in daylight. Secondly, our approach requires that one plant per frame for accurate analysis because it has been trained on a dataset with one plant per image. For real-time deployment, plant instance detection from an image followed by its extraction may be required, which can increase processing overheads.

\section{Conclusion}
In this paper, a novel deep learning-based pipeline for plant water stress (water deficiency) phenotyping has been proposed and validated via a detailed study on water stress identification from Chickpea plant shoot images. The pipeline consists of four main stages - image sequence input, data augmentation (and input processing), Convolutional Neural Network - Long Short Term Memory (CNN-LSTM) network, and water stress prediction. There are no publicly available datasets of pulses plant shoot images, specifically Chickpea plant, so a new dataset of two varieties of chickpea shoot images under different water stress conditions has been considered. The authors ensured high-quality training data by taking adequate measures during data acquisition and applying data augmentation techniques like Gaussian noise augmentation before neural processing. The proposed pipeline employs a CNN-LSTM to learn visual spatio-temporal patterns and use them to classify water stress. This work demonstrates that temporal analysis of chickpea plant shoot images outperforms the best time-invariant (or only spatial) analysis by nearly $14\%$ for both chickpea varieties. Further, the experimental results show that the temporal approach is independent of the underlying CNN feature extractor. This study also illustrates the robustness of our proposed CNN-LSTM model to noise. Across both species, the average model accuracy dipped by less than  2.5\%,  with a  small standard deviation, thereby ensuring high and consistent classification capabilities even in noisy conditions. Moreover, the Grad-CAM visualizations explain and validate the use of chickpea plant shoot images to detect water stress. The ablation study further reveals that the proposed CNN-LSTM model, consequently the proposed deep learning pipeline, performs equitably on both water stress-sensitive species JG-62 and stress-resistant Pusa-372. Finally, the results of all four experiments in this paper validate the stress-sensitive nature of JG-62 and the stress-tolerant nature of Pusa-372. The findings in this paper demonstrate the potential of the proposed technique for real-time applications like plant stress monitoring and intelligent irrigation. However, this technique is not without its caveats. The proposed method has been validated on a controlled dataset while ensuring a high degree of resemblance to real-world conditions and data. Techniques like noise removal and plant shoot segmentation may be required while dealing with real-world data, which may, in turn, increase computational overheads. Nevertheless, we believe that the proposed deep learning pipeline will form the basis for future work in this domain. We encourage researchers to validate our work on their datasets and build upon this pipeline. Our future works will also be focused on proposing new components for our deep learning pipeline that will make it more robust and take it closer to real-world deployment. We are also experimenting with lightweight models that will be less-compute intensive. 

\ifCLASSOPTIONcaptionsoff
  \newpage
\fi

\bibliographystyle{IEEEtran}
\bibliography{paper.bib}

\begin{IEEEbiography}[{\includegraphics[width=1in,height=1.25in,clip,keepaspectratio]{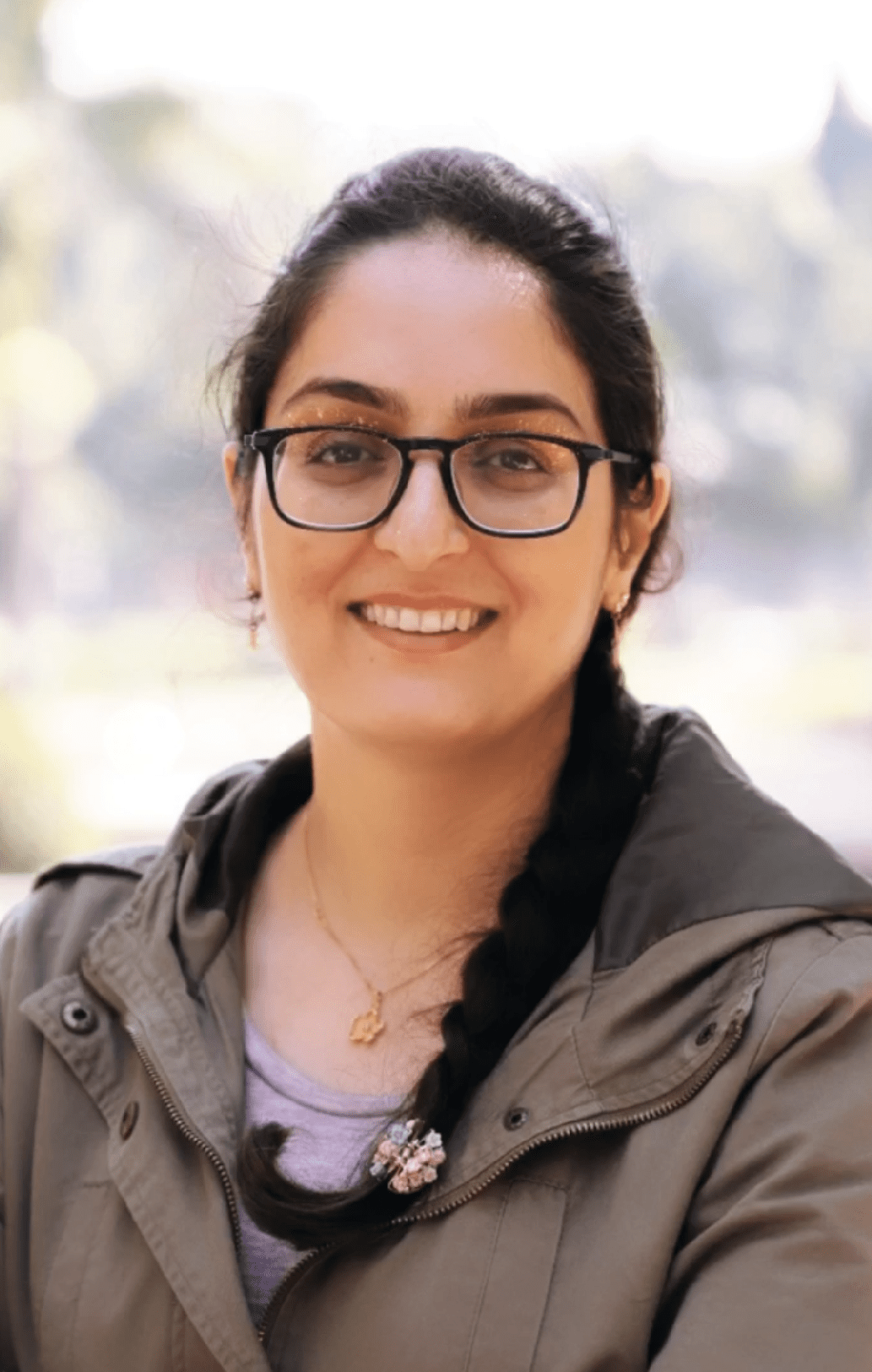}}]{Shiva Azimi} received her B.Tech. degree in electronics and communication engineering from Islamic Azad University, Iran, in 2008, and the M.Tech. degree in image processing from Islamic Azad university, Najafabad, Iran, in 2013. She is currently pursuing the Ph.D. degree with the Department of Electrical Engineering, Indian Institute of Technology Delhi, India. Her research interests include plant phenotyping, computer vision, image processing, and machine learning.
\end{IEEEbiography}

\begin{IEEEbiography}[{\includegraphics[width=1in,height=1.25in,clip,keepaspectratio]{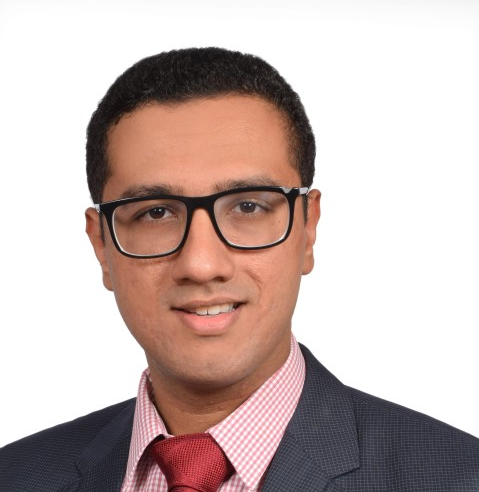}}]{Rohan Wadhawan} (Member, IEEE) is a Research Affiliate with the Department of Electrical Engineering, IIT Delhi, India. He completed a B.E. in Computer Engineering at Netaji Subhas Institute of Technology, University of Delhi, India, in August 2020. His research interests include computer vision, image processing, machine learning, neural processing, generative learning, multimodal learning, and affective computing. The detailed biographical information can be found at: \url{https://www.rohanwadhawan.com}
\end{IEEEbiography}

\begin{IEEEbiography}[{\includegraphics[width=1in,height=1.25in,clip,keepaspectratio]{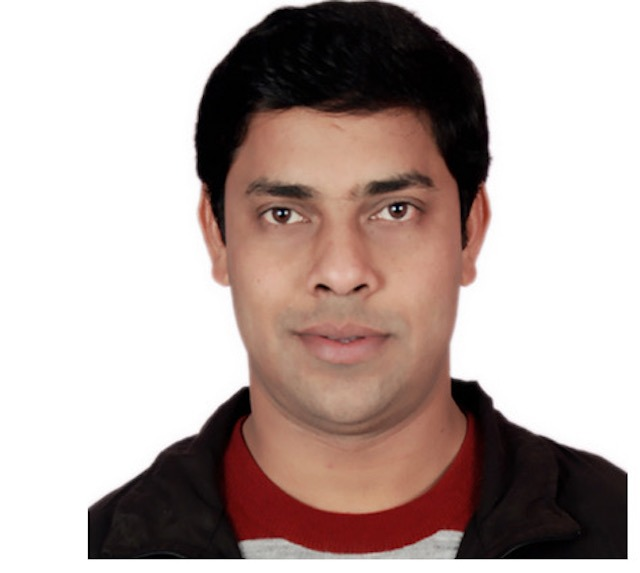}}]{Tapan K. Gandhi} (Senior Member, IEEE) received the B.Sc. degree in physics, the M.Sc. degree in electronics, the M.Tech. degree in bio electronics, and the Ph.D. degree in biomedical engineering from IIT Delhi, in 2001, 2003, 2006, and 2011, respectively. He is currently an Associate Professor with the Department of Electrical Engineering, IIT Delhi. Prior to joining IIT Delhi, as a Faculty Member, he was a Postdoctoral Fellow with the Massachusetts Institute of Technology (MIT), Cambridge, USA, for three years. His research interests include cognitive computation, artificial intelligence, medical instrumentation, biomedical signal and image processing, and assistive technology.
\end{IEEEbiography}

\end{document}